\documentclass{article}
\usepackage[preprint]{nested_flows_style}
\usepackage{placeins,longtable,multirow,xcolor}
\usepackage[colorlinks=true,linkcolor=accent,citecolor=accent,urlcolor=accent]{hyperref}
\usepackage[capitalise,noabbrev,nameinlink]{cleveref}

\hypersetup{pdfauthor={Mohammad Mahdi Derakhshani, Pedro M. P. Curvo, Gertjan J. Burghouts, Jan-Willem van de Meent, Cees G. M. Snoek},pdftitle={LiFT: Loop Flow Transformers -- Loop in Depth, Flow in Time}}
\title{LiFT: Loop Flow Transformers\\[3pt]
\large Loop in Depth, Flow in Time}
\author{Mohammad Mahdi Derakhshani$^{1,4,*}$\quad Pedro M.\ P.\ Curvo$^{1,4,*}$\quad Gertjan J.\ Burghouts$^{3,\dagger}$\\
\bfseries Jan-Willem van de Meent$^{2,4,\dagger}$\quad Cees G.\ M.\ Snoek$^{1,4,\dagger}$\\
\normalfont $^{1}$VISLab\quad $^{2}$AMLab\quad $^{3}$TNO, Intelligent Imaging\quad $^{4}$University of Amsterdam\\
\normalfont\texttt{\{m.m.derakhshani,\,p.m.pombeirocurvo\}@uva.nl}\\[2pt]
\normalfont\small $^{*}$Equal co-authorship; random order.\quad $^{\dagger}$Equal supervision.}
\begin{document}
\raggedbottom
\setlength{\textfloatsep}{12pt plus 2pt minus 2pt}
\setlength{\intextsep}{10pt plus 2pt minus 2pt}
\maketitle
\begin{abstract}
We introduce Loop Flow Transformers (LiFT), a family of looped generative models that scales computation by repeatedly applying a shared Diffusion Transformer (DiT) core, with only light changes to the standard architecture.
Rather than asking every recurrent step for the final prediction, LiFT trains each step with a single regression target: a point on a straight path from the model's initial estimate to the flow-matching target.
Because we index these targets by a continuous depth coordinate, a trained model can loop far beyond its training depth with no retraining, early exits, or other modifications.
In our experiments, these longer rollouts improve generation, so inference computation can grow without adding parameters.
On ImageNet at $256\times256$, LiFT-L/2 achieves an FID 3.34 points lower than our dense DiT-XL/2 baseline while using approximately 60\% fewer parameters, 32\% fewer training FLOPs, and 52\% fewer inference FLOPs.
\end{abstract}

\begin{center}
\small
\begin{tabular}{@{}c@{\hspace{2.5em}}c@{\hspace{2.5em}}c@{}}
  \raisebox{-.2\height}{\includegraphics[height=1.4em]{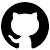}}
  \enspace Coming Soon
  & \raisebox{-.2\height}{\includegraphics[height=1.4em]{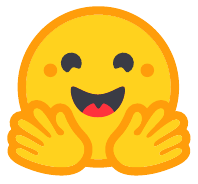}}
  \enspace Coming Soon
  & \raisebox{-.2\height}{\includegraphics[height=1.4em]{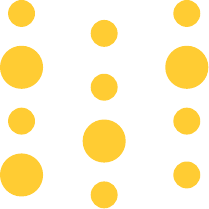}}
  \enspace Coming Soon
\end{tabular}
\end{center}

\begin{figure}[H]
\centering
\includegraphics[width=\linewidth]{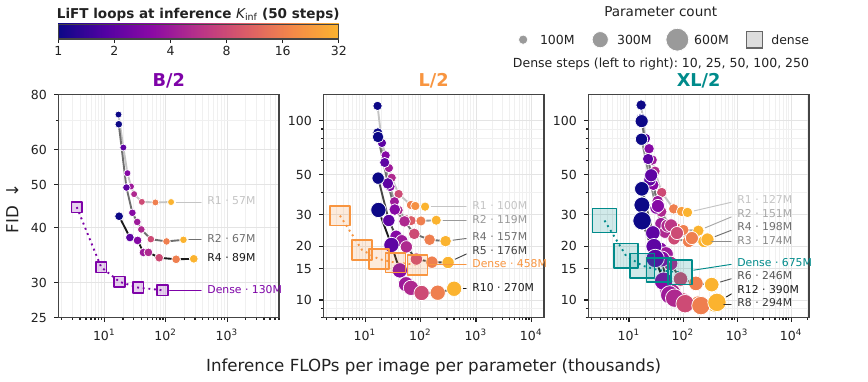}
\caption{\textbf{Looping a smaller model beats larger dense DiTs at L/2 and
XL/2 scale.} FID against inference FLOPs per parameter for LiFT checkpoints at
50 integration steps (solid; colour encodes inference depth, marker area
parameter count) and for dense DiTs as the number of steps varies (dotted). Only at B/2,
the dense model remains stronger.}
\label{fig:parameter-compute}
\end{figure}

\clearpage
\section{Introduction}
\label{sec:introduction}

Diffusion Transformers (DiTs) have become a dominant architecture for
generative modeling, with increasingly large models driving advances
across image, video, and scientific
domains~\citep{peebles2023,esser2024sd3,polyak2024moviegen,nvidia2025cosmos}.
Much of this progress, however, has come from increasing parameter
count, so that generation quality remains closely tied to model size.

Looped transformers offer an alternative: by repeatedly applying a shared
set of layers, they increase the network's effective depth while keeping
the number of trainable parameters
fixed~\citep{dehghani2019universal,geiping2025recurrent}.
In flow-based generation~\citep{lipman2023,liu2023rectified,albergo2023interpolants},
this opens a second way to spend inference compute. A flow model
generates an image by following a learned velocity field from noise to
data, and the usual way to add compute is to take more integration
steps, each requiring one network evaluation.
Each step, however, can be no better than the network's velocity estimate,
so quality saturates once integration error is no longer the
bottleneck~\citep{xu2023restart,ma2025inference}. Looping instead spends
additional compute within each step to refine the velocity estimate, so steps
and loops act as complementary axes of inference
compute~\citep{goyal2026elt}. Steps follow the flow more closely; loops
estimate it more accurately. In principle, a trained looped model can
therefore buy better predictions at test time by running more loops.

In practice, however, extra loops pay off only when two conditions are met.
A model must be able to handle more loops than it was trained with, and
the added compute must yield better quality than a dense model given the
same compute. Results on both points have so far been mixed. Without
dedicated training, a looped diffusion transformer produces coherent
images only at its training depth~\citep{goyal2026elt}. Intermediate
supervision can make looped models usable across depths, but loops beyond the training depth soon hurt
rather than help: on ImageNet, FID worsens as soon as inference exceeds
the training depth, and on video, quality peaks at 1.5 times the training
depth~\citep{goyal2026elt}. In these methods every intermediate prediction
is trained toward the complete output~\citep{shin2025deepflow,goyal2026elt},
so no stage is given a distinct role. On the second condition, a looped DiT-B/4
trained with MeanFlow gave no advantage over a dense model
at equal compute, and extra sampling steps paid off more than extra
loops~\citep{chai2026loopdit}.

We address both conditions with the
\textbf{Loop Flow Transformer (LiFT)}, which, rather than asking every
loop for the complete output, gives each loop its own target.
Within each network evaluation, LiFT draws a straight path from the
model's initial velocity estimate to its training target and tells each
loop how far along this path it is. Each loop is supervised at the
matching point, so successive loops carry out successive stages of a
single correction, and the last loop is trained on the target itself.
Because progress along the path is continuous, the loop count only sets
how finely the path is traversed, much as the number of integration steps
does for generative time; both can be chosen after training. LiFT thus
separates computational depth from generative time:
\par\smallskip
{\centering
\textbf{\emph{\textquotedblleft{}Loop in depth, flow in time.\textquotedblright{}}}
\par}
\smallskip
\noindent
Our results on ImageNet $256\times256$ (\cref{fig:parameter-compute}) show
that at the DiT-L/2 and DiT-XL/2 scales~\citep{peebles2023}, LiFT
outperforms larger dense DiTs with fewer parameters. The figure separates
the number of integration steps $T$ from the number of loops used in
training, $K_{\mathrm{train}}$, and at inference, $K_{\mathrm{inf}}$. Dense
DiTs level off as $T$ grows to 250 steps. A LiFT checkpoint trained
at a fixed $K_{\mathrm{train}}$ trails a dense DiT that executes the same
number of blocks, but keeps improving as $K_{\mathrm{inf}}$ grows until it
reaches FIDs the dense models do not attain, at lower
inference cost. At DiT-B/2, the dense model remains stronger.

\begin{bluebox}{Contributions}
\begin{enumerate}
  \setlength{\itemsep}{3pt}
  \setlength{\parskip}{0pt}

  \item \textbf{A trajectory objective for recurrent generation.}
  We introduce LiFT, which gives each loop its own target on a path
  from the model's initial estimate to the training target, and trains the
  last loop on the standard flow-matching objective.

  \item \textbf{Useful recurrence far beyond training depth.}
  Without retraining, an XL/2 checkpoint trained with two loops
  reduces FID from \textbf{17.07 to 9.31} at sixteen loops, eight times
  its training loop count.

  \item \textbf{Better generation with fewer parameters and lower inference cost.}
  On ImageNet $256\times256$, LiFT-L/2 lowers FID by 3.34 relative to
  our dense DiT-XL/2 baseline, with about 60\% fewer parameters, 52\% fewer
  inference FLOPs, and the same training-token budget.

\end{enumerate}
\end{bluebox}

\clearpage
\section{Background}
\label{sec:background}

\subsection{Flow Matching}
\label{sec:bg-flow-matching}

Let $\rho_0$ denote a tractable base distribution, typically
$\mathcal N(0,I)$, and let $\rho_1=p_{\mathrm{data}}$ denote the target
data distribution. Flow matching learns a time-dependent velocity field
that transports samples from $\rho_0$ to $\rho_1$~\citep{lipman2023}.
A simple construction, also underlying rectified flow and a special case
of stochastic interpolants~\citep{liu2023rectified,albergo2023interpolants},
starts from independent draws $x_0\sim\rho_0$ and $x_1\sim\rho_1$ and
defines the linear conditional path and its velocity as
\begin{equation}
  x_t=(1-t)x_0+t x_1,\qquad
  u^\star=\frac{\mathrm d x_t}{\mathrm dt}=x_1-x_0,
  \quad t\in[0,1].
  \label{eq:bg-interpolant}
\end{equation}
The marginal distribution $\rho_t$ of $x_t$ interpolates between the
base and data distributions. Because the corresponding sample-specific
velocity target is available directly, training can regress it without
simulating an ODE. With
$t\sim\mathcal U(0,1)$ and $d$ output coordinates, the conditional
flow-matching objective is
\begin{equation}
  \mathcal L_{\mathrm{FM}}(\theta)
  =\mathbb E_{x_0,x_1,t}\!\left[
    \frac1d\|v_\theta(x_t,t)-u^\star\|_2^2\right].
  \label{eq:bg-flow-loss}
\end{equation}
Its unrestricted regression optimum is the conditional mean
$v^\star(x,t)=\mathbb E[u^\star\mid x_t=x,t]$, the marginal velocity
associated with the path. Additional conditioning, such as a class label,
can be included in both the predictor and this conditional expectation.
At inference, the learned field defines a neural
ODE~\citep{chen2018neural}: generation starts from $X_0\sim\rho_0$ and
numerically integrates $\mathrm dX_t/\mathrm dt=v_\theta(X_t,t)$ to
$t=1$, yielding an approximation to $\rho_1$. We write $T$ for the number of integration steps; with ordinary Euler
integration, each step calls the learned field once.

\subsection{Diffusion Transformers}
\label{sec:bg-dit}

Diffusion Transformers (DiTs) parameterize diffusion or flow-based generative
models with a transformer backbone operating over image or latent
patches~\citep{peebles2023}. The input is first embedded into a sequence of
tokens and then processed by a stack of attention and feed-forward blocks.
Generative time $t$ and, when available, auxiliary information such as a
class label condition these blocks through mechanisms such as adaptive
normalization and gated residual branches. A final prediction head maps
the resulting token representations back to the input geometry to produce
the quantity required by the generative objective. In our setting, this
quantity is the flow-matching velocity $v_\theta(x_t,t)$.

\subsection{Looped Transformers and Computational Depth}
\label{sec:bg-looped}

A conventional transformer assigns distinct parameters to each block. A looped transformer instead applies one shared block, or group of blocks, repeatedly, so executed depth grows with the number of passes while the parameter count stays fixed~\citep{dehghani2019universal}. This distinction can be
expressed through a prelude--core--coda decomposition, following
\citet{geiping2025recurrent} and the architectural formulation of
\citet{chen2026scalingexponents}. For an input $x$ and conditioning $c$,
\begin{equation}
  h_0=P_\theta(x,c),\qquad
  h_k=R_\theta\!\left(\phi(h_{k-1},h_0),c\right),\qquad
  u_k=\mathcal D_\theta(h_k,c).
  \label{eq:bg-recurrence}
\end{equation}
The \emph{prelude} $P_\theta$ embeds the input once, producing the initial state $h_0$. The \emph{core} $R_\theta$ is applied $K$ times with the same weights; the \emph{boundary operator} $\phi$ links successive passes and may normalize the state and reinject $h_0$ so that every pass retains access to the input. The \emph{coda} $\mathcal D_\theta$ maps any state to a prediction through its own transformer blocks, if any, and a prediction head, so $u_0$ is the prediction from the prelude alone and $u_K$ the prediction after $K$ passes. The core may additionally receive an embedding of its pass index; \cref{sec:lift} builds on this. In our setting $x=x_t$, $c$
encodes $t$ and any class label, and $v_\theta(x_t,t)=u_K$.

With $L_P$, $L_R$, and $L_D$ transformer blocks in the prelude, core, and coda, respectively, weight sharing separates the number of unique blocks from the number executed to produce the final prediction:
\[
  L_{\mathrm{unique}}=L_P+L_R+L_D,\qquad
  L_{\mathrm{executed}}=L_P+K L_R+L_D.
\]
These counts exclude the embedding and prediction heads and assume that the coda is evaluated only at the final state. This separation also underlies iso-depth scaling studies, which compare repeated computation with distinct layers at matched executed depth~\citep{schwethelm2026recurrence}. The loop count $K$ therefore controls computation within a single prediction, independently of any generative time coordinate. Training unrolls a chosen number of passes; at inference, their number can change
without changing the trained parameters. Whether additional passes
improve the prediction depends on the learned recurrence.

\paragraph{Parameter and compute budgets.}
Following \citet{hoffmann2022chinchilla}, let $N$ denote the number of
trainable parameters, $D$ the number of training tokens processed, and
$C_{\mathrm{train}}$ the total training compute in FLOPs. For image
models, tokens are patches of an image or its latent representation.
With $n$ tokens per example, $U$ optimizer updates, and global batch
size $B_{\mathrm{batch}}$, the training-data budget is
\[
  D=U B_{\mathrm{batch}}\,n.
\]
Repeated presentations of an image contribute to this count, whereas
recurrent passes over the same tokens increase computation without
increasing $D$ or adding parameters. Equal parameter or data budgets
therefore need not imply equal training compute.

We separately measure inference compute per generated image as
\[
  C_{\mathrm{infer}}=T\,C_{\mathrm{forward}}(K),
\]
where $T$ is the number of integration steps and
$C_{\mathrm{forward}}(K)$ is the cost of one velocity prediction with $K$
loops. These costs exclude tokenizer decoding.

\section{LiFT}
\label{sec:lift}

We introduce the Loop Flow Transformer (LiFT), a flow-based generative model that increases computational depth by repeatedly applying a shared transformer core. To train this recurrence, we construct a continuous reference path from the model's initial velocity estimate to the flow-matching target. During training, we randomly sample intermediate depth coordinates $s$, condition each core application on its coordinate, and supervise the resulting predictions at the corresponding points along the path. Each loop thus receives a distinct correction task, while the final prediction retains the ordinary flow-matching objective.

\subsection{Method}
\label{sec:lift_method}

\begin{figure}[!htb]
\centering
\includegraphics[width=\linewidth]{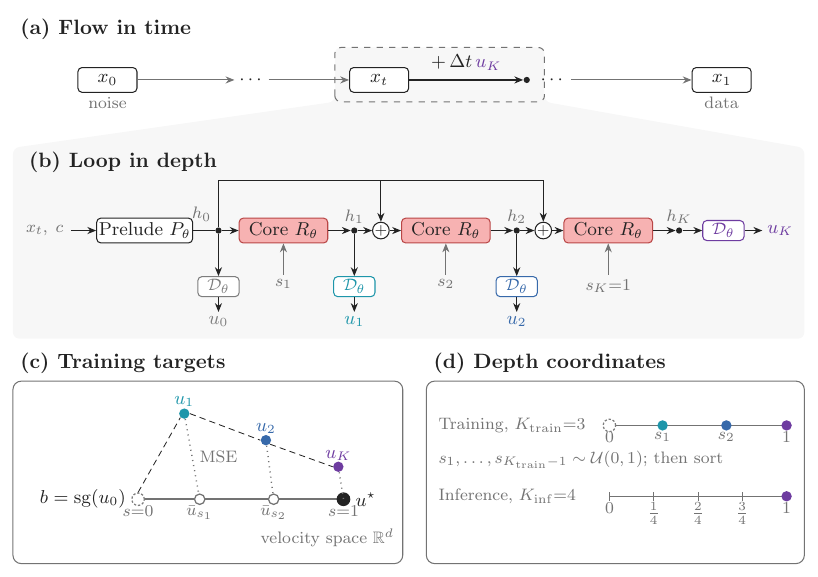}
\caption{\textbf{Loop in depth, flow in time.} (a)~The sampler advances
through generative time, calling the network once per step. (b)~Within
each call, a shared core is applied $K$ times, each time at its own depth
coordinate $s_k$, and a shared readout maps every state to a prediction
$u_k$. (c)~In training, each $u_k$ is regressed onto its point
$\bar u_{s_k}$ on the path from $b=\operatorname{sg}(u_0)$ to $u^\star$.
(d)~Training samples the interior coordinates at random; inference uses a
uniform grid. Colours mark the same loop across panels; normalization and
the first reinjection of $h_0$ are omitted for clarity.}
\label{fig:lift-overview}
\end{figure}

For a training example $(x_0,x_1,t)$, let $x_t$ and $u^\star$ denote the
interpolated state and conditional velocity target defined in
\cref{eq:bg-interpolant}. We introduce a continuous computational-depth coordinate
$s\in[0,1]$ to describe progress within a single velocity evaluation,
keeping the noisy input, generative time, and any class condition fixed.
\Cref{fig:lift-overview} expands one such evaluation, the dashed sampler
step in panel~(a), into the recurrence of panel~(b): starting from a
prelude state $h_0$, $K$ applications of the shared core produce states
$h_1,\ldots,h_K$, and a shared readout maps each state $h_k$ to a
prediction $u_k$ at depth $s_k$. The loop count is chosen separately for the
two phases: training unrolls $K_{\mathrm{train}}$ loops, whereas inference
may use a different depth $K_{\mathrm{inf}}$; plain $K$ appears only in
statements that hold for both. We suppress the
fixed outer conditioning here and specify the recurrence in
\cref{sec:lift_architecture}.

\begin{bluebox}{A reference path in prediction space.}
The initial prediction defines an anchor $b=\operatorname{sg}(u_0)$,
where $\operatorname{sg}$ denotes stop-gradient. Because it is computed
before the recurrent rollout, this estimate depends on neither the
sampled grid nor the loop count. We connect it to the sampled
flow-matching target through the linear path
\begin{equation}
    \bar u_s=(1-s)b+s u^\star,
    \qquad s\in[0,1].
    \label{eq:lift_reference}
\end{equation}
\end{bluebox}
For a fixed initial predictor $b=b(z)$, where $z$ collects the noisy
input, generative time, and any class condition, the reference path has
conditional mean
$\mathbb E[\bar u_s\mid z]=(1-s)b(z)+s\,v^\star(z)$, where
$v^\star(z)=\mathbb E[u^\star\mid z]$ is the marginal velocity of
\cref{sec:bg-flow-matching}. Its squared-loss regression optimum
therefore progresses linearly from the initial estimate toward the
marginal flow field, reaching the ordinary flow-matching optimum at
$s=1$. Each depth coordinate thus defines a prediction task along the
same correction path; \cref{sec:app-statistical} gives the
assumptions and derivation.

\begin{bluebox}{Random depth coordinates.}
To train the recurrence across this interval, we independently sample
$K_{\mathrm{train}}-1$ coordinates from $\mathcal U(0,1)$ for each example at every
update (\cref{fig:lift-overview}d, top). Sorting these draws and appending the endpoints gives
\begin{equation}
    s_0=0<s_1<\cdots<s_{K_{\mathrm{train}}-1}<s_{K_{\mathrm{train}}}=1.
    \label{eq:lift_depth_grid}
\end{equation}
\end{bluebox}
We then unroll the same $K_{\mathrm{train}}$ core applications at these coordinates,
supervising each readout against its corresponding reference value, as
in \cref{fig:lift-overview}c.
Varying the sampled grid changes both the intermediate prediction tasks
and the recurrent histories through which they are reached, while the
rollout length stays fixed. Every grid retains the endpoint $s_{K_{\mathrm{train}}}=1$,
so the final readout always addresses the complete flow-matching task.

\begin{bluebox}{Trajectory supervision over $K_{\mathrm{train}}$ loops.}
Writing $\mathcal S=(s_0,\ldots,s_{K_{\mathrm{train}}})$ for the sampled grid, we train
the recurrent predictions with
\begin{equation}
    \mathcal L_{\mathrm{LiFT}}(\theta)
    =
    \mathbb E_{x_0,x_1,t,\mathcal S}
    \left[
        \frac{1}{K_{\mathrm{train}}d}
        \sum_{k=1}^{K_{\mathrm{train}}}
        \left\|
            u_k-\bigl((1-s_k)b+s_k u^\star\bigr)
        \right\|_2^2
    \right],
    \label{eq:lift_trajectory_loss}
\end{equation}
\end{bluebox}
Here $d$ denotes the prediction dimension, and the loss averages
uniformly over the $K_{\mathrm{train}}$ recurrent readouts, with any class condition
included in the expectation. At the endpoint
$s_{K_{\mathrm{train}}}=1$, the reference recovers the conditional flow-matching
target, while each intermediate coordinate specifies how much
of the correction remains:
\[
u^\star-\bar u_s=(1-s)(u^\star-b).
\]
The objective therefore distributes the prediction task across
depth, supervising a progression toward the endpoint rather
than asking every readout to recover it immediately.

Although the anchor $b$ is recomputed on every forward pass, it is detached
when forming the targets, whereas gradients through the predictions
propagate through the full recurrent rollout, including the prelude.

\paragraph{Supervising the initial estimate.}
The trajectory objective trains the recurrent predictions but does not
directly require $u_0$ to approximate the target velocity. As the shared
parameters change, its magnitude can drift, allowing the anchor
contribution $(1-s)b$ to dominate the intermediate targets and leaving
the core to learn an unnecessarily large correction. To encourage a
useful starting estimate, we add an auxiliary regression loss:
\begin{equation}
    \begin{aligned}
        \mathcal L_{\mathrm{train}}(\theta)
        &=
        \mathcal L_{\mathrm{LiFT}}(\theta)
        +\lambda\,\mathcal L_{\mathrm{prelude}}(\theta),\\
        \mathcal L_{\mathrm{prelude}}(\theta)
        &=
        \mathbb E_{x_0,x_1,t}
        \left[
            \frac{1}{d}\|u_0-u^\star\|_2^2
        \right].
    \end{aligned}
    \label{eq:lift_training_loss}
\end{equation}
The coefficient $\lambda$ controls supervision of the initial prediction
and is applied once outside the $1/K_{\mathrm{train}}$ average, making its weight
independent of the training loop count. Whereas this auxiliary term
differentiates through $u_0$, the same prediction remains detached when
constructing the trajectory targets. Consequently, at $K_{\mathrm{train}}=1$, training
combines the ordinary flow-matching objective with the separate prelude
loss. \Cref{alg:lift-training} in \cref{app:pseudocode} summarizes the
procedure, with experimental settings reported in
\cref{app:experimental-details}.

\paragraph{Inference with a chosen depth budget.}
At each integration step, LiFT initializes a fresh prelude state
and applies the shared core at
$s_k=k/K_{\mathrm{inf}}$ (\cref{fig:lift-overview}d, bottom). Changing the loop budget therefore
changes the computational grid while preserving its endpoint
$s=1$. The final readout at this endpoint supplies the velocity
to the outer sampler of \cref{fig:lift-overview}a, while intermediate readouts are needed
only during training. With Euler integration, the update is
\begin{equation}
x_{t+\Delta t}
=
x_t+\Delta t\,u_\theta^{(K_{\mathrm{inf}})}(x_t,t),
\label{eq:lift_sampling}
\end{equation}
where $u_\theta^{(K_{\mathrm{inf}})}(x_t,t)$ denotes the final readout
$u_{K_{\mathrm{inf}}}$, and $\Delta t=1/T$
for $T$ uniform steps.

\begin{samepage}
\subsection{Architecture}
\label{sec:lift_architecture}

LiFT partitions a DiT backbone into a prelude $P_\theta$, a shared
recurrent core $R_\theta$, and a prediction-only coda $\mathcal D_\theta$ that
includes the velocity readout. The prelude embeds the input patches
and processes them once per integration step, while the core reuses
the same transformer blocks at every inner step. The fixed outer
conditioning $c=e_t(t)+e_y(y)$ sums embeddings of the generative time $t$
and the class label $y$, omitting $e_y$ when no label is available, and
the computation is
\begin{equation}
    \begin{aligned}
        h_0 &= P_\theta(x_t,c),\\
        h_k &= R_\theta\!\left(
            \operatorname{RMSNorm}(h_{k-1})+h_0,\,
            c+e_s(s_k)
        \right),
        && k=1,\ldots,K,\\
        u_k &= \mathcal D_\theta(h_k,c),
        && k=0,\ldots,K.
    \end{aligned}
    \label{eq:lift_architecture}
\end{equation}
\end{samepage}
The boundary operation normalizes the previous state and reinjects $h_0$
(the $\oplus$ nodes in \cref{fig:lift-overview}b),
preserving access to the input representation throughout the rollout.
In addition to this input reinjection, we adapt the shared transformation
to its position along the reference path by conditioning the core on
$e_s(s_k)$ through adaptive normalization and residual gates,
following timestep-modulated looped transformers~\citep{xu2025timestep}.
This depth-dependent conditioning is confined to the core, whereas the
prelude and coda use only the fixed outer condition $c$. Within the core,
however, the depth embedding supplies only the current coordinate $s_k$;
the preceding coordinate and interval length remain implicit in the
recurrent history rather than appearing as explicit inputs.

The coda $\mathcal D_\theta$ maps each hidden state to a velocity
prediction using parameters shared across depths. Because it serves
only as a prediction branch, it does not modify the state passed to
the next core application.

With $L_P$, $L_R$, and $L_D$ transformer blocks in the prelude, core,
and coda, LiFT contains $L_P+L_R+L_D$ unique blocks but executes
$L_P+K_{\mathrm{inf}} L_R+L_D$ blocks at inference. Because trajectory training
also evaluates the coda at the initial state and every intermediate
state, its forward pass instead executes
$L_P+K_{\mathrm{train}} L_R+(K_{\mathrm{train}}+1)L_D$
transformer blocks. These counts exclude embeddings and prediction
heads; \cref{app:trajectory_compute} provides analytic FLOP accounting
that includes these operations and backward computation.

\section{Experiments}
\label{sec:experiments}

LiFT makes recurrent depth adjustable at inference, allowing a
fixed set of parameters to support different computational budgets.
We evaluate how this flexibility affects generation quality on
class-conditional ImageNet, addressing four questions:

\begin{enumerate}
  \setlength{\itemsep}{2pt}
  \setlength{\parskip}{0pt}
  \item[\textbf{Q1.}] Does LiFT improve beyond its training depth, and
  overtake dense DiTs?
  \item[\textbf{Q2.}] How does LiFT compare with larger dense
  models at comparable compute?
  \item[\textbf{Q3.}] How should compute be split between recurrent
  depth and integration steps?
  \item[\textbf{Q4.}] Which evaluated settings best fit a compute or
  parameter budget?
\end{enumerate}

\subsection{Experimental setup}
\label{sec:experimental_setup}

We evaluate class-conditional generation on ImageNet-1k at
$256\times256$ resolution, using SD-VAE representations of shape
$4\times32\times32$.

\paragraph{Models.}
Our dense DiT baselines and LiFT use the same transformer blocks,
adapted from SpeedrunDiT~\citep{bhanded2025speedrundit}. These blocks use RMSNorm,
per-head RMS normalization of attention queries and keys, and
two-dimensional rotary positional embeddings, while retaining
GELU MLPs and adaLN-zero-style conditioning with gated residual
branches. Sharing these modifications across model families controls
for the underlying block design.
We use the B/2, L/2, and XL/2 backbone dimensions, each with
$2\times2$ latent patches.

In the main comparisons, LiFT uses four prelude blocks ($L_P=4$),
a shared recurrent core, and no transformer blocks in the coda
($L_D=0$). The coda therefore consists only of a shared prediction
head, which normalizes and conditionally modulates the tokens before
projecting and unpatchifying the velocity prediction. During training,
this head reads both the prelude and recurrent states; during sampling,
it reads only the final state.
\cref{tab:model_configurations} lists representative configurations;
\cref{tab:all_model_configurations} in
\cref{app:data_and_models} gives the complete list.

\paragraph{Training.}
We fix
the training-data budget at $D=3.2768\times10^{10}$ latent-patch
tokens per model across all main experiments. With 256 tokens
per image, this corresponds to 500,000 optimizer updates at a
global batch size of 256, independently of recurrent depth.
All models share the data pipeline and optimization settings,
using AdamW with a learning rate of $10^{-4}$. Dense baselines
are trained with conditional flow matching, while LiFT uses
trajectory supervision with an auxiliary prelude-loss weight
of $\lambda=0.01$.

Within each backbone scale, we vary the shared-core size $L_R$ and training depth
$K_{\mathrm{train}}$ while keeping the executed transformer depth
$4+K_{\mathrm{train}}L_R$ equal to the dense reference: 12 blocks for
B/2, 24 for L/2, and 28 for XL/2. With the same training-token budget,
this approximately matches transformer training computation while
changing the balance between unique parameters and repeated
computation. Our FLOP estimates additionally account for conditioning
and intermediate readouts. With no transformer coda, their additional
cost is less than $0.2\%$ of the dense reference across these
configurations; \cref{app:trajectory_compute} derives the
comparison and gives the numerical budgets.

\paragraph{Evaluation.}
We sample with exponential-moving-average weights and Euler integration
without classifier-free guidance, pairing the initial noise and class
labels across configurations. For each setting, we generate 50,000
images and compute FID, spatial FID (sFID), and Inception Score (IS)
from those same images using the official ADM evaluator. FID serves
as the primary metric in the main comparisons. \Cref{app:full_results}
lists every evaluated setting with its inference cost, FID, sFID, and IS,
and \cref{tab:all_model_configurations} collects parameter counts and
training costs, so every number quoted in this section can be traced to
a table. Full optimization settings, evaluation details, and compute
accounting are provided in \cref{app:experimental-details}.

\begin{table}[t]
\centering
\small
\setlength{\tabcolsep}{5pt}
\caption{\textbf{Same executed depth, far fewer parameters.}
Representative LiFT configurations and their dense references; $L_R$ is the shared-core block count and $L_{\mathrm{exec}}^{\mathrm{train}}$ the executed training depth, including the $L_P=4$ prelude blocks, and $N$ the parameter count. \cref{tab:all_model_configurations} lists all configurations.}
\label{tab:model_configurations}
\begin{tabular}{lrrrr}
\toprule
Model & $L_R$ & $K_{\mathrm{train}}$ & $L_{\mathrm{exec}}^{\mathrm{train}}$ & $N$ (M) \\
\midrule
DiT-B/2 & -- & -- & 12 & 130.30 \\
DiT-L/2 & -- & -- & 24 & 457.83 \\
DiT-XL/2 & -- & -- & 28 & 674.82 \\
\midrule
LiFT-B/2-R1 & 1 & 8 & 12 & 56.69 \\
LiFT-B/2-R4 & 4 & 2 & 12 & 88.58 \\
LiFT-L/2-R5 & 5 & 4 & 24 & 175.79 \\
LiFT-L/2-R10 & 10 & 2 & 24 & 270.24 \\
LiFT-XL/2-R8 & 8 & 3 & 28 & 293.96 \\
LiFT-XL/2-R12 & 12 & 2 & 28 & 389.58 \\
\bottomrule
\end{tabular}
\end{table}

\FloatBarrier
\subsection{Q1: Generalization beyond training depth}
\label{sec:depth_generalization}
\label{sec:matched_training_compute}

\begin{bluebox}{Question:}
Does LiFT improve beyond its training depth, and overtake dense DiTs?
\end{bluebox}

\begin{figure}[htbp]
\centering
\includegraphics[width=\linewidth]{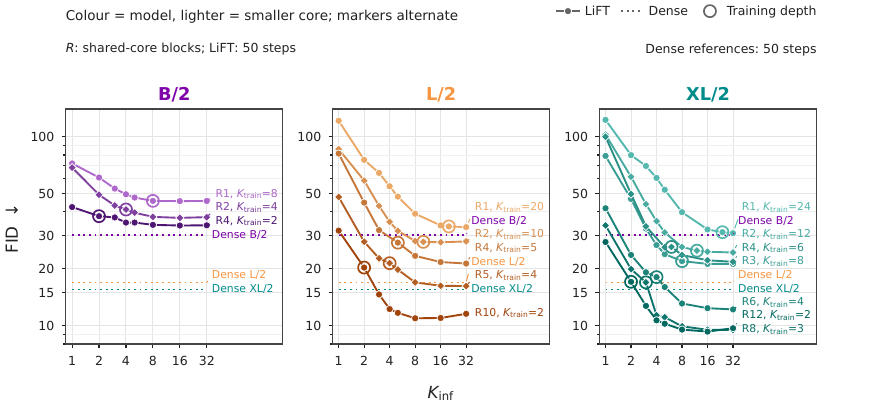}
\caption{\textbf{LiFT improves beyond its training depth, and its larger cores overtake dense DiTs.} FID versus inference depth at 50 integration steps; rings mark training depth and dotted lines the dense DiTs. At L/2 and XL/2, the largest cores keep improving well past training depth and overtake the dense model of their scale, whereas single-block cores plateau. Training FLOPs match within each scale.}

\label{fig:depth-generalization}
\end{figure}

To assess depth generalization, we vary $K_{\mathrm{inf}}$ for each checkpoint with the sampler fixed at 50 integration steps. \Cref{fig:depth-generalization} shows FID against inference depth, with training depths marked and dense B/2, L/2, and XL/2 drawn as horizontal references.

Increasing inference depth substantially reduces FID without further
optimization. L/2 R10 improves from 20.30 at its two-loop training
depth to 10.95 at eight inference loops; XL/2 R12 from 17.07 to 9.31 at sixteen inference loops; XL/2 R8 from 16.95 at three training loops to 9.50 at thirty-two inference loops. Checkpoints trained at larger depth follow the same trend with smaller gains (L/2 R5: 21.42 at four training loops to 16.21 at thirty-two inference loops), while the single-block cores plateau near their training depth, consistent with the limited capacity of a single shared block~\citep{goyal2026elt}. The curves of the largest cores flatten after eight to sixteen inference loops, with L/2 R10 and XL/2 R12 rising slightly at thirty-two loops.
\Cref{tab:depth_extrapolation_summary} summarizes the best depth for representative checkpoints.

Inference depth alone also carries several checkpoints past the dense references. L/2 R10 overtakes both dense L/2 and dense XL/2 at three inference loops, and XL/2 R12 overtakes dense XL/2 at three as well. XL/2 R8 and R6 follow at four and eight inference loops, whereas L/2 R5 draws level with dense L/2 at eight inference loops and overtakes it at sixteen. Because $T$ is fixed, most of these crossings cost more inference FLOPs than the dense model they beat. The exception is L/2 R10 at three inference loops, which outperforms dense XL/2 (FID 14.64 versus 15.54) with slightly less inference compute (11.43 versus 11.86 TFLOPs) and 60\% fewer parameters.

\begin{table}[t]
\centering
\small
\setlength{\tabcolsep}{5pt}
\caption{\textbf{The best inference depth is often far beyond the training depth.} FID at training depth $K_{\mathrm{train}}$ and best FID over inference depth $K_{\mathrm{inf}}$, at ratio $r=K_{\mathrm{inf}}/K_{\mathrm{train}}$;
\cref{tab:all_depth_extrapolation} lists all checkpoints.}
\label{tab:depth_extrapolation_summary}
\begin{tabular}{lrrrrr}
\toprule
Model & $K_{\rm train}$ & FID at $K_{\rm train}$ & Best $K_{\rm inf}$ & Best FID & $r$ \\
\midrule
B/2 R4 & 2 & 37.95 & 16 & 33.87 & $8\times$ \\
L/2 R10 & 2 & 20.30 & 8 & 10.95 & $4\times$ \\
XL/2 R12 & 2 & 17.07 & 16 & 9.31 & $8\times$ \\
XL/2 R8 & 3 & 16.95 & 32 & 9.50 & $10.67\times$ \\
L/2 R5 & 4 & 21.42 & 32 & 16.21 & $8\times$ \\
B/2 R1 & 8 & 45.70 & 16 & 45.58 & $2\times$ \\
\bottomrule
\end{tabular}
\end{table}

\FloatBarrier
\begin{finding}
Training depth is not a ceiling: a checkpoint trained with few loops can
be run with many more, turning a small model into a deeper one at
inference.
\end{finding}

\subsection{Q2: Generation quality at comparable compute}
\label{sec:compute_comparison}
\label{sec:inference_efficiency}

\begin{figure}[t]
\centering
\includegraphics[width=\linewidth]{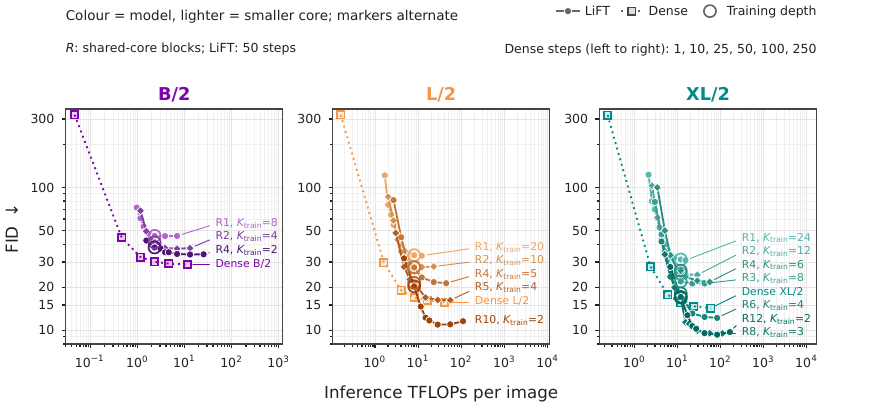}
\caption{\textbf{Looping more reaches lower FID than sampling more for L/2
and XL/2.} FID versus inference cost: LiFT varies loops at 50 integration
steps (solid), and dense DiTs vary the number of steps (dotted); rings
mark training depth. Training FLOPs match within each model scale only.}
\label{fig:fixed_nfe_inference_efficiency}
\end{figure}

\begin{bluebox}{Question:}
How does LiFT compare with larger dense models at comparable compute?
\end{bluebox}
In Q1 we fixed $T$, so most crossings of a dense reference cost more inference FLOPs than the dense model they beat. Here we compare on cost,
$C_{\mathrm{infer}}=T\,C_{\mathrm{forward}}(K_{\mathrm{inf}})$: LiFT keeps 50 integration steps and varies $K_{\mathrm{inf}}$, while the dense references vary $T$
(\cref{fig:fixed_nfe_inference_efficiency}).

At its training depth, each LiFT checkpoint executes as many transformer blocks as its dense reference and therefore matches its inference cost, yet every evaluated checkpoint attains a higher FID than the same-scale dense model at 50 steps. For example, L/2 R10 reaches FID 20.30 at its two training loops, compared with 16.94 for dense L/2 at nearly identical cost. This gap is expected: at matched executed depth, a looped model replaces unique blocks with shared ones, and iso-depth scaling laws for looped language models find that a shared recurrence is worth only about half of a unique block~\citep{schwethelm2026recurrence}.
The advantage of LiFT therefore emerges not at the training operating
point but once test-time compute is reallocated, whether by extending
recurrent depth beyond training or, as Q3 will show, by trading integration
steps for loops.

\Cref{fig:fixed_nfe_inference_efficiency} shows that, once depth is
extended, the outcome depends on scale. At B/2, the dense model remains stronger at every
evaluated cost: the best LiFT B/2 result, FID 33.87, does not reach the
32.50 that dense B/2 attains with 25 integration steps at 1.150 TFLOPs.
We attribute this gap to capacity: with at most 89M parameters for the same
32.8B training tokens, the B/2 cores appear too small for the data budget,
so additional inference loops cannot compensate.
At L/2 and XL/2, by contrast, recurrence yields operating points that
improve on the dense reference while costing less. LiFT L/2 R10 with
four inference loops attains FID 12.25 at 14.79 TFLOPs per image, whereas dense
L/2 needs 16.14 TFLOPs and 100 steps to reach 16.17. Similarly,
LiFT XL/2 R8 with four inference loops reaches FID 11.36 at 15.25 TFLOPs, compared
with 14.69 at 23.72 TFLOPs for dense XL/2. Because each pair shares a
backbone scale, these gains hold under comparable training FLOPs and
identical training-token budgets, while LiFT uses 41\% and 56\% fewer
parameters, respectively.

The same sweeps also permit a comparison across scales, although
training budgets then differ. With eight inference loops, LiFT L/2 R10 attains
FID 10.95 at 28.24 TFLOPs, whereas dense XL/2 reaches 14.30 only with
250 integration steps at 59.31 TFLOPs. LiFT therefore lowers FID by 3.34
points while using approximately 60\% fewer parameters and 52\% fewer
inference FLOPs. Moreover, LiFT reaches this quality with about 32\% less
training computation (61.98 versus 91.10 EFLOPs).

\begin{finding}
A smaller recurrent model can stand in for a larger dense one. At equal
depth it trails, as its fewer parameters predict, but, once the shared core has enough parameters, looping more at inference closes and then reverses the gap.
\end{finding}

\subsection{Q3: Allocating computation across depth and time}
\label{sec:compute_allocation}

\begin{figure}[t]
\centering
\includegraphics[width=\linewidth]{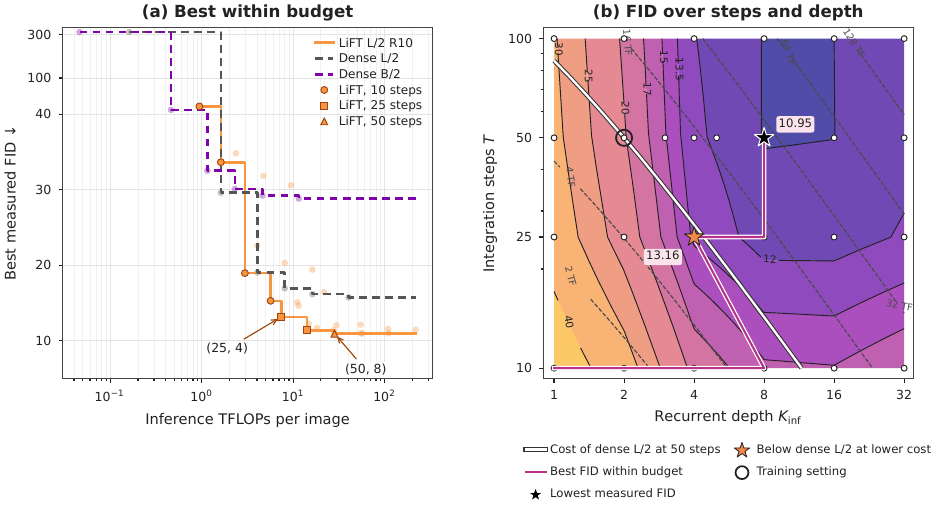}
\caption{\textbf{Balancing inference loops and integration steps gives the best quality per budget.} (a) Lowest measured FID within each inference budget for LiFT L/2 R10, which varies both, and for dense DiTs, which vary integration steps; labels give $(T,K_{\mathrm{inf}})$. (b) FID across inference loops and integration steps, interpolated between the measured dots; dashed lines connect allocations of equal cost, the white line marks the cost of dense L/2 at 50 steps (FID 16.94), and the pink path traces the best allocation as the budget grows. Dense L/2 matches LiFT's training FLOPs.}
\label{fig:inference-efficiency}
\label{fig:joint_budget_frontier}
\end{figure}

\begin{bluebox}{Question:}
How should compute be split between recurrent depth and integration steps?
\end{bluebox}
The comparisons in Q2 fix LiFT at 50 steps, leaving
open how a given budget should be divided between recurrent depth and
integration steps. We therefore evaluate LiFT L/2 R10 over the joint
grid $T\in\{10,25,50,100\}$ and
$K_{\mathrm{inf}}\in\{1,2,4,8,16,32\}$. For each inference budget,
\cref{fig:inference-efficiency}(a) reports the lowest FID among the
settings that fit within it, alongside the dense envelopes, whereas
\cref{fig:inference-efficiency}(b) maps FID across the full grid, so that
allocations of equal cost can be compared directly.

Allowing both budgets to vary sharpens the comparison with dense L/2.
With 25 integration steps and four inference loops, LiFT L/2 R10 achieves FID
13.16 at 7.397 TFLOPs per image, compared with 16.94 at 8.069 TFLOPs for
dense L/2 with 50 steps, and thus improves generation with
approximately 41\% fewer parameters and 8\% less inference computation.
In \cref{fig:inference-efficiency}(b), this setting lies on the cheaper
side of the white curve that marks the cost of dense L/2.
The same setting also improves on the checkpoint's own training
configuration: relative to two training loops and 50 steps, it reduces FID
from 20.30 to 13.16 while lowering cost from 8.070 to 7.397 TFLOPs. In
this regime, recurrent refinement is a more efficient use of inference
compute than additional integration steps.

Recurrence is not, however, always the better destination for
additional compute. At sixteen inference loops and ten integration steps, the
same model obtains FID 15.09 at 11.027 TFLOPs, which is both worse and
more expensive than the four-loop, 25-step setting. Here, shifting
computation toward integration steps improves the trade-off, so the
preferable direction of reallocation depends on the starting
configuration.

At larger budgets, the FID surface also flattens. With eight inference loops,
doubling the number of steps from 50 to 100 leaves FID nearly unchanged at
10.95 and 10.96, whereas extending depth to thirty-two inference loops raises FID at
both budgets. Even at this plateau, the best measured setting, FID 10.95
at 28.24 TFLOPs, remains well below the 15.74 that dense L/2 reaches
with 250 steps at 40.35 TFLOPs. \Cref{app:budget_envelopes}
reports the selected operating points, and \cref{app:full_results} the
complete FID, sFID, and IS sweeps.

\begin{finding}
Inference quality depends on balancing $K_{\mathrm{inf}}$ against
$T$: neither more loops nor more integration steps alone
yields the best result at a given budget.
\end{finding}

\subsection{Q4: Model selection under resource budgets}
\label{sec:resource_budget_selection}

\begin{bluebox}{Question:}
Which evaluated settings best fit a compute or parameter budget?
\end{bluebox}

\begin{figure}[t]
\centering
\includegraphics[width=\linewidth]{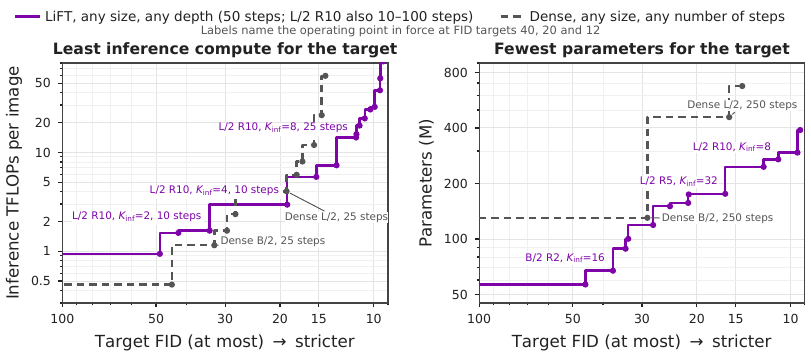}
\caption{\textbf{LiFT needs fewer parameters for every quality target, and
less compute for strict ones.} Least inference compute (left) and fewest
parameters (right) with which any evaluated LiFT or dense setting reaches
each target FID; settings span scales with different training FLOPs.}
\label{fig:target-budgets}
\end{figure}

Q3 varies the inference settings of a fixed checkpoint, but deployment
also requires choosing the checkpoint itself. To support this choice,
\cref{fig:target-budgets} reports, for each target FID, the least
inference compute and the fewest parameters with which any evaluated
checkpoint reaches it. Conversely, a fixed budget on either resource
identifies the best measured FID it can support. The two resources need
not favor the same model, however. For example, at a target FID of 20,
minimizing compute selects LiFT L/2 R10 with ten integration steps and
four inference loops (2.96 TFLOPs per image), whereas minimizing parameters selects
the smaller LiFT L/2 R5 (176M parameters).

\begin{finding}
When memory is the constraint, LiFT reaches any quality target with fewer
parameters than a dense DiT, and strict targets with less compute as well.
\end{finding}

\subsection{Qualitative comparisons}
\label{sec:qualitative}

\begin{figure}[t]
    \centering
    \includegraphics[width=\linewidth]{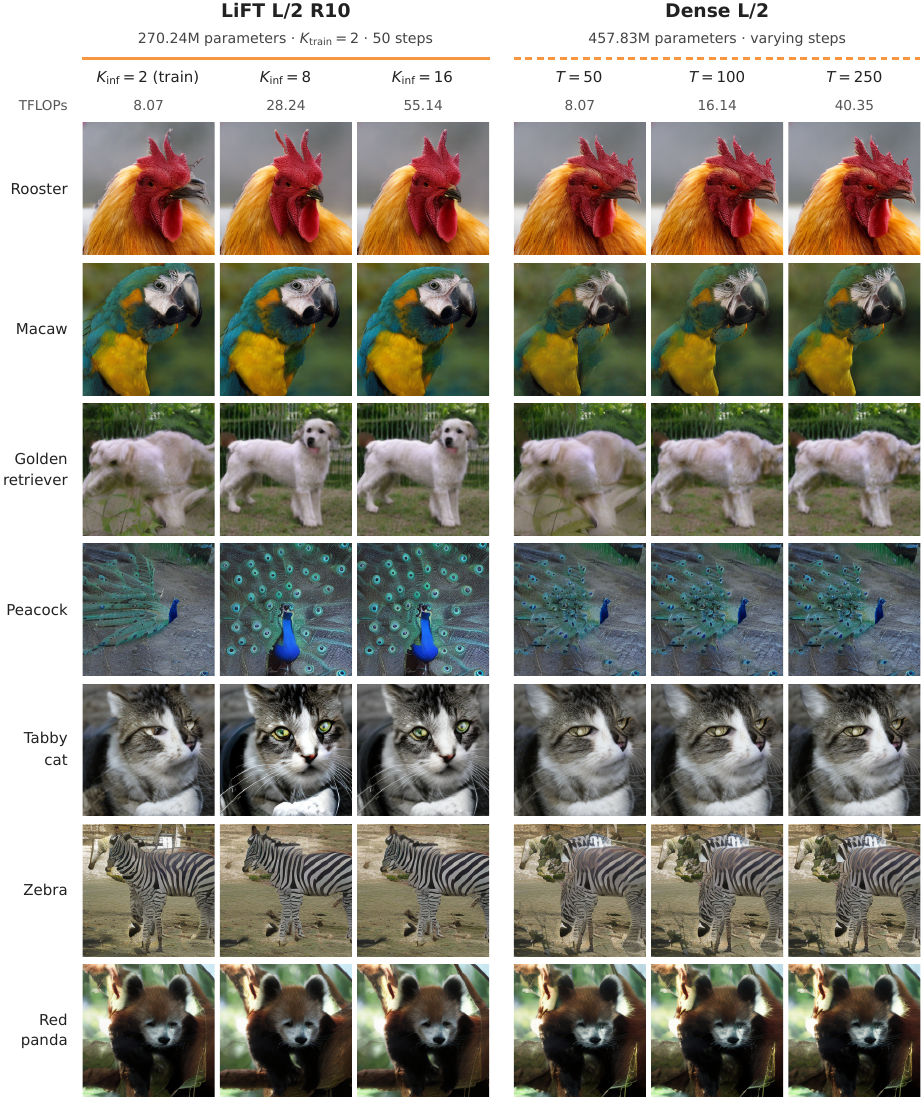}
    \caption{\textbf{More loops refine samples from the same noise.} Selected
    paired samples from LiFT L/2 R10 at 50 integration steps and dense L/2 at
    varying numbers of steps; each row shares class and initial noise, and
    headers give inference TFLOPs per image. Compare the second column with
    the last: at eight inference loops, LiFT uses 30\% less compute than dense
    L/2 at 250 steps and reaches a lower FID (10.95 versus 15.74).}
    \label{fig:qualitative_depth}
\end{figure}

\Cref{fig:qualitative_depth} complements these comparisons with paired
generations from LiFT L/2 R10 and dense L/2.

\section{Related Work}
\label{sec:related_work}

\paragraph{Recurrent computation and depth generalization.}
Reusing a shared transformation separates the number of trainable
parameters from the depth of the executed computation, as in Universal
Transformers~\citep{dehghani2019universal}. The ability to benefit from
additional iterations at inference also predates recent looped
transformers: on algorithmic reasoning tasks,
\citet{schwarzschild2021algorithm} demonstrate generalization beyond a
fixed training depth, while \citet{bansal2022extrapolation} investigate
training strategies that sustain useful computation over longer rollouts.
Recurrent language models extend this direction to larger scales,
including training over distributions of recurrence
counts~\citep{geiping2025recurrent}. Together, these works establish
recurrent depth as a resource for inference-time computation; LiFT
investigates how intermediate generative predictions should be supervised
to make use of that resource.

\paragraph{Depth conditioning and computational trajectories.}
To adapt a shared transformation across recurrent applications,
\citet{xu2025timestep} condition it on computational depth.
LoopFormer~\citep{jeddi2026loopformer} extends this perspective to
normalized trajectories, using position and step-size conditioning
to train shorter computations toward a full-depth solution. Whereas
LoopFormer aligns trajectory endpoints across computational budgets,
LiFT assigns distinct targets to intermediate velocity
predictions within each rollout. Its normalized coordinate specifies
the corresponding fraction of the correction toward the flow-matching
target, while training keeps the number of recurrent applications
fixed and randomizes their intermediate coordinates.

\paragraph{Recurrence in generative transformers.}
Within generative networks, DiT-Air~\citep{chen2025ditair} investigates
layer sharing in diffusion transformers, while \citet{ai2026depth}
examine recurrent layer arrangements inside flow-matching
text-to-speech models.
LoopDiT~\citep{chai2026loopdit} provides a close architectural comparison:
both methods combine a prelude, a shared recurrent core with
depth-dependent conditioning, and an output stage. Within this shared
layout, LoopDiT trains its final prediction with MeanFlow, whereas
LiFT supervises the intermediate readouts of a flow-matching velocity
predictor. The two methods also allocate shorter computations differently:
LoopDiT retains a prefix of its training conditioning schedule, while
LiFT changes the grid over the full interval $[0,1]$, so every budget
reaches $s=1$.

\paragraph{Intermediate supervision for generative prediction.}
Beyond the sharing layout, the training objective determines what
intermediate predictions are asked to recover.
DeepFlow~\citep{shin2025deepflow} studies intermediate flow-matching
supervision and velocity alignment, while
R-MDM~\citep{carballo2026recursive} examines how the generative loss
should be distributed across recurrent outputs.
ELT~\citep{goyal2026elt} supplements supervision at sampled intermediate
exits with self-distillation from a detached full-depth prediction and
also reports depth extrapolation in UCF-101 video generation.
LiFT organizes these prediction tasks along the reference
$\bar u_s=(1-s)\operatorname{sg}(u_0)+s\,u^\star$, assigning each readout
a fraction of the correction toward the
flow-matching target. Unlike a full-depth teacher used in
self-distillation, the detached initial prediction defines the starting
point from which this correction is measured. The resulting
depth-indexed targets define different prediction tasks along the
trajectory, while the final readout retains the complete endpoint task.

\paragraph{Computational depth and generative time.}
The distinction between computation within an evaluation and progression
through the generative process also appears in continuous-depth models.
$\mathrm{ODE}_t(\mathrm{ODE}_l)$~\citep{gudovskiy2026ode} explicitly
separates sampling time from network length and uses length consistency
to relate predictions produced with different computational budgets.
LiFT instead uses its continuous coordinate to define a reference
trajectory in prediction space. Its core remains a discrete,
depth-conditioned recurrence: additional loops change the rollout rather
than refine a prescribed numerical solver for an inner ODE.
A separate distinction concerns whether recurrent state is reused
within a velocity evaluation or carried across the generative trajectory.
RIN~\citep{jabri2023rin} and Thinking with Looped
Flows~\citep{suleymanzade2026thinking} carry representations as the
generative input changes, whereas LiFT initializes its recurrent state
separately at each integration step. Flow Reasoning
Models~\citep{helbling2026flow} provide a closer connection to refinement
at fixed generative input: they recurrently condition on previous clean
predictions and use Fixed-Point Forcing to construct detached,
rollout-derived conditioning states. LiFT instead carries hidden
features, differentiates through the recurrent rollout, and assigns
depth-dependent velocity targets to its readouts.

\paragraph{Other forms of recursion and equilibrium.}
Recursive Flow Matching~\citep{huang2026recursive} relates rescaled
generative trajectories through velocity and consistency supervision;
its recursion connects transport problems rather than intermediate
predictions within one velocity evaluation. Fixed Point Diffusion
Models~\citep{bai2024fixed}, Generative Equilibrium
Transformers~\citep{geng2023equilibrium}, and
CoFRe~\citep{miele2026fixed} instead define repeated computation through
a hidden-state equilibrium. These perspectives are complementary to
LiFT's finite, depth-conditioned rollout, whose reference path specifies
how its predictions should progress without imposing a fixed-point
equation on the hidden states.

\section{Conclusion}
LiFT gives each training loop of a recurrent DiT its own target, a point on a
reference path from the model's initial velocity estimate to the
flow-matching target. With this supervision, recurrent depth behaves as a
second inference-time axis: a fixed checkpoint keeps improving well beyond
its training depth, and the best quality at a given budget comes from
balancing inference loops against integration steps rather than maximizing either.
Although LiFT trails a dense DiT of the same executed depth at its
training depth, reallocating inference compute toward recurrence lets its
larger-core L/2 and XL/2 checkpoints outperform larger dense DiTs with
fewer parameters and less inference compute.

\paragraph{Implications.}
By making recurrent depth selectable at inference, LiFT allows a fixed
checkpoint to adapt its computation to the available budget.
Because the shared core adds executed depth without adding parameters,
this flexibility could be particularly useful when parameter storage
is more restrictive than computation, including memory-constrained
edge devices. It also motivates further study of budget-aware
generation in video and generative dynamics models.

\paragraph{Limitations.}
Our evaluation concerns class-conditional image generation at
$256\times256$ resolution without classifier-free guidance.
Whether the same depth-scaling behavior extends to other
modalities or guided sampling remains open. Within this setting, the
advantage of LiFT depends on scale and operating point: at B/2, the
dense model remains stronger at every evaluated cost, and at its
training depth every LiFT checkpoint trails its same-scale dense
reference. Each result also comes from a single training run and sample
set. The deployment implications require hardware evaluation, since we
report parameter counts and analytic FLOP estimates rather than
device-level latency, energy consumption, or peak memory. Finally,
inference depths are prescribed, and the joint allocation study covers
one checkpoint over a finite grid; adaptive depth selection and broader
optimization of inner and outer compute budgets remain for future study.

\section*{Acknowledgements}

This project was supported by the ELLIS Unit Amsterdam and by SURF
through access to the Snellius supercomputer. Cees G. M. Snoek is (partially) funded by the Horizon Europe project ELLIOT (GA No. 101214398).

\clearpage

\bibliographystyle{unsrtnat}
\bibliography{references}

\clearpage
\appendix
\section*{Appendix index}
\pdfbookmark[0]{Appendix index}{appendix-index}

\begingroup
\small
\setlength{\tabcolsep}{0pt}
\renewcommand{\arraystretch}{1.25}
\newcommand{\appentry}[2]{%
  \hyperref[#1]{\ref*{#1}\quad #2} & \pageref{#1}\\}
\newcommand{\appgroup}[2]{%
  \multicolumn{2}{@{}l@{}}{\textbf{\hyperref[#1]{\ref*{#1}\quad #2}}}\\[2pt]}
\begin{tabular}{@{}p{.93\linewidth}r@{}}
 & Page\\
\appgroup{app:full_results}{Complete and extended experimental results}
\appentry{app:all_evaluations}{Complete numerical results corresponding to the main figures}
\appentry{app:depth_generalization}{Depth-generalization matrices}
\appentry{app:joint_budget}{Joint allocation of inference computation}
\noalign{\vskip 8pt}
\appgroup{app:ablations}{Ablations and diagnostics}
\appentry{app:depth_coordinates}{Random versus fixed depth coordinates}
\appentry{app:prelude_loss}{Prelude-loss / $\lambda$ ablation}
\appentry{app:prediction_trajectories}{Prediction trajectories}
\appentry{app:grid_consistency}{Computational-grid consistency}
\noalign{\vskip 8pt}
\appgroup{sec:app-derivations}{Analysis of the LiFT objective}
\appentry{sec:app-statistical}{Conditional regression result}
\appentry{sec:app-gradient}{Gradient flow with a detached anchor}
\appentry{sec:app-action}{Euclidean minimum action}
\appentry{sec:app-geometric}{Gaussian, Fisher, and KL interpretations}
\noalign{\vskip 8pt}
\appgroup{app:experimental-details}{Reproducibility and implementation}
\appentry{app:data_and_models}{Data and architecture}
\appentry{app:training_protocol}{Training protocol}
\appentry{app:pseudocode}{Training and sampling pseudocode}
\appentry{app:evaluation_protocol}{Evaluation protocol}
\appentry{app:compute_accounting}{Compute accounting}
\end{tabular}
\endgroup

\clearpage
\section{Complete and extended experimental results}
\label{app:full_results}

This appendix reports every measured setting behind the main figures,
including small regressions, and extends the analysis of depth
extrapolation and of the allocation of inference compute. Each operating
point is evaluated once on 50,000 generated images, so small FID
differences, such as those near the plateaus, have not been tested for
statistical significance. \Cref{app:experimental-details} gives the
architecture, training, and evaluation details.

\subsection{Complete numerical results corresponding to the main figures}
\label{app:all_evaluations}

\Cref{tab:all_evaluations} lists the dense step sweeps and LiFT depth
sweeps plotted in the main comparisons, with sampling costs computed as
in \cref{app:compute_accounting}. Training depths, parameter counts, and
training costs are collected in \cref{tab:all_model_configurations}.

\begingroup
\small
\setlength{\tabcolsep}{4pt}
\begin{longtable}{lrrrrrr}
\caption{Complete numerical results corresponding to the main figures.
Each row reports FID, sFID, and IS from the same 50,000 generated images.
R denotes the number of blocks in the shared core; a dash denotes a dense model.}
\label{tab:all_evaluations}\\
\toprule
Model & $K_{\mathrm{inf}}$ & $T$ & TFLOPs/image & FID & sFID & IS \\
\midrule
\endfirsthead
\multicolumn{7}{l}{\tablename~\thetable{} (continued)}\\
\toprule
Model & $K_{\mathrm{inf}}$ & $T$ & TFLOPs/image & FID & sFID & IS \\
\midrule
\endhead
\bottomrule
\endfoot
 & -- & 1 & 0.046 & 321.17 & 408.50 & 2.04 \\*
 & -- & 10 & 0.460 & 44.49 & 13.18 & 45.92 \\*
\raisebox{-.5\baselineskip}[0pt][0pt]{Dense B/2} & -- & 25 & 1.150 & 32.50 & 8.92 & 53.42 \\*
 & -- & 50 & 2.300 & 30.10 & 7.84 & 54.64 \\*
 & -- & 100 & 4.600 & 29.22 & 7.39 & 55.11 \\*
 & -- & 250 & 11.501 & 28.82 & 7.19 & 54.86 \\
\midrule
 & -- & 1 & 0.161 & 321.15 & 410.26 & 2.02 \\*
 & -- & 10 & 1.614 & 29.59 & 10.99 & 68.33 \\*
\raisebox{-.5\baselineskip}[0pt][0pt]{Dense L/2} & -- & 25 & 4.035 & 19.05 & 6.90 & 82.98 \\*
 & -- & 50 & 8.069 & 16.94 & 5.98 & 86.15 \\*
 & -- & 100 & 16.139 & 16.17 & 5.67 & 86.99 \\*
 & -- & 250 & 40.347 & 15.74 & 5.52 & 87.61 \\
\midrule
 & -- & 1 & 0.237 & 319.32 & 409.10 & 2.03 \\*
 & -- & 10 & 2.372 & 27.80 & 9.86 & 72.85 \\*
\raisebox{-.5\baselineskip}[0pt][0pt]{Dense XL/2} & -- & 25 & 5.931 & 17.68 & 6.57 & 89.16 \\*
 & -- & 50 & 11.862 & 15.54 & 5.81 & 92.56 \\*
 & -- & 100 & 23.723 & 14.69 & 5.52 & 93.38 \\*
 & -- & 250 & 59.308 & 14.30 & 5.43 & 93.83 \\
\midrule
 & 1 &  & 0.959 & 72.11 & 10.90 & 23.70 \\*
 & 2 &  & 1.151 & 60.71 & 11.14 & 28.14 \\*
 & 3 &  & 1.342 & 53.09 & 9.84 & 31.52 \\*
\raisebox{-.5\baselineskip}[0pt][0pt]{LiFT B/2 R1} & 4 & \raisebox{-.5\baselineskip}[0pt][0pt]{50} & 1.534 & 49.54 & 9.24 & 32.91 \\*
 & 5 &  & 1.726 & 47.60 & 8.91 & 33.52 \\*
 & 8 &  & 2.301 & 45.70 & 8.32 & 33.79 \\*
 & 16 &  & 3.834 & 45.58 & 7.99 & 33.61 \\*
 & 32 &  & 6.901 & 45.67 & 7.92 & 33.46 \\
\midrule
 & 1 &  & 1.151 & 68.55 & 11.82 & 24.45 \\*
 & 2 &  & 1.534 & 49.27 & 9.63 & 33.50 \\*
 & 3 &  & 1.917 & 43.18 & 8.49 & 36.79 \\*
\raisebox{-.5\baselineskip}[0pt][0pt]{LiFT B/2 R2} & 4 & \raisebox{-.5\baselineskip}[0pt][0pt]{50} & 2.301 & 41.16 & 7.83 & 37.60 \\*
 & 5 &  & 2.684 & 39.58 & 7.06 & 38.22 \\*
 & 8 &  & 3.834 & 37.62 & 6.33 & 39.29 \\*
 & 16 &  & 6.900 & 37.18 & 6.15 & 39.34 \\*
 & 32 &  & 13.033 & 37.50 & 6.21 & 39.16 \\
\midrule
 & 1 &  & 1.534 & 42.40 & 7.50 & 39.06 \\*
 & 2 &  & 2.300 & 37.95 & 6.67 & 40.19 \\*
 & 3 &  & 3.067 & 37.37 & 6.41 & 40.61 \\*
\raisebox{-.5\baselineskip}[0pt][0pt]{LiFT B/2 R4} & 4 & \raisebox{-.5\baselineskip}[0pt][0pt]{50} & 3.833 & 35.10 & 5.66 & 41.95 \\*
 & 5 &  & 4.600 & 35.11 & 5.68 & 42.00 \\*
 & 8 &  & 6.900 & 34.15 & 5.41 & 42.71 \\*
 & 16 &  & 13.032 & 33.87 & 5.34 & 43.06 \\*
 & 32 &  & 25.296 & 33.96 & 5.34 & 43.21 \\
\midrule
 & 1 &  & 1.682 & 121.21 & 104.66 & 7.27 \\*
 & 2 &  & 2.018 & 75.29 & 10.05 & 22.40 \\*
 & 3 &  & 2.355 & 64.22 & 10.11 & 26.97 \\*
 & 4 &  & 2.691 & 54.58 & 9.60 & 31.10 \\*
LiFT L/2 R1 & 5 & 50 & 3.027 & 48.16 & 9.24 & 34.91 \\*
 & 8 &  & 4.036 & 39.08 & 8.42 & 40.80 \\*
 & 16 &  & 6.727 & 33.92 & 7.42 & 44.82 \\*
 & 20 &  & 8.072 & 33.50 & 7.28 & 45.18 \\*
 & 32 &  & 12.108 & 33.21 & 7.17 & 45.60 \\
\midrule
 & 1 &  & 2.018 & 85.76 & 10.23 & 17.47 \\*
 & 2 &  & 2.691 & 58.55 & 8.86 & 28.08 \\*
 & 3 &  & 3.363 & 42.97 & 7.98 & 38.21 \\*
 & 4 &  & 4.036 & 35.41 & 7.55 & 44.81 \\*
LiFT L/2 R2 & 5 & 50 & 4.708 & 31.74 & 7.34 & 48.56 \\*
 & 8 &  & 6.726 & 28.04 & 6.96 & 53.47 \\*
 & 10 &  & 8.071 & 27.75 & 6.89 & 53.92 \\*
 & 16 &  & 12.106 & 27.63 & 6.74 & 53.71 \\*
 & 32 &  & 22.865 & 27.90 & 6.71 & 53.21 \\
\midrule
 & 1 &  & 2.691 & 81.29 & 10.49 & 19.10 \\*
 & 2 &  & 4.036 & 44.75 & 9.04 & 36.04 \\*
 & 3 &  & 5.380 & 32.01 & 6.69 & 48.01 \\*
LiFT L/2 R4 & 5 & 50 & 8.070 & 27.50 & 6.36 & 53.61 \\*
 & 8 &  & 12.104 & 23.41 & 5.03 & 59.29 \\*
 & 16 &  & 22.863 & 21.71 & 4.50 & 61.80 \\*
 & 32 &  & 44.380 & 21.32 & 4.36 & 62.56 \\
\midrule
 & 1 &  & 3.027 & 47.91 & 8.58 & 35.52 \\*
 & 2 &  & 4.708 & 27.80 & 7.22 & 56.95 \\*
 & 3 &  & 6.389 & 22.71 & 6.58 & 65.41 \\*
\raisebox{-.5\baselineskip}[0pt][0pt]{LiFT L/2 R5} & 4 & \raisebox{-.5\baselineskip}[0pt][0pt]{50} & 8.070 & 21.42 & 6.22 & 67.99 \\*
 & 5 &  & 9.751 & 19.81 & 5.47 & 70.97 \\*
 & 8 &  & 14.794 & 16.94 & 4.73 & 77.30 \\*
 & 16 &  & 28.242 & 16.27 & 4.55 & 79.02 \\*
 & 32 &  & 55.138 & 16.21 & 4.52 & 79.18 \\
\midrule
 & 1 &  & 4.708 & 31.83 & 6.22 & 51.22 \\*
 & 2 &  & 8.070 & 20.30 & 6.34 & 71.80 \\*
 & 3 &  & 11.431 & 14.64 & 4.22 & 84.12 \\*
\raisebox{-.5\baselineskip}[0pt][0pt]{LiFT L/2 R10} & 4 & \raisebox{-.5\baselineskip}[0pt][0pt]{50} & 14.793 & 12.25 & 4.12 & 92.01 \\*
 & 5 &  & 18.155 & 11.70 & 4.33 & 93.53 \\*
 & 8 &  & 28.241 & 10.95 & 4.74 & 97.17 \\*
 & 16 &  & 55.136 & 10.99 & 4.62 & 96.87 \\*
 & 32 &  & 108.926 & 11.57 & 4.07 & 93.75 \\
\midrule
 & 1 &  & 2.119 & 122.52 & 18.87 & 11.80 \\*
 & 2 &  & 2.543 & 79.77 & 12.75 & 18.44 \\*
 & 3 &  & 2.967 & 69.91 & 10.71 & 23.90 \\*
 & 4 &  & 3.391 & 60.46 & 11.55 & 27.93 \\*
LiFT XL/2 R1 & 5 & 50 & 3.814 & 52.36 & 10.87 & 31.92 \\*
 & 8 &  & 5.086 & 39.83 & 8.97 & 40.41 \\*
 & 16 &  & 8.476 & 32.28 & 7.29 & 47.04 \\*
 & 24 &  & 11.866 & 31.23 & 7.04 & 47.95 \\*
 & 32 &  & 15.256 & 30.85 & 6.93 & 48.22 \\
\midrule
 & 1 &  & 2.543 & 102.40 & 44.09 & 12.42 \\*
 & 2 &  & 3.390 & 61.26 & 8.36 & 26.55 \\*
 & 3 &  & 4.238 & 43.92 & 7.90 & 37.17 \\*
 & 4 &  & 5.085 & 35.72 & 7.88 & 44.80 \\*
LiFT XL/2 R2 & 5 & 50 & 5.932 & 31.14 & 7.52 & 50.13 \\*
 & 8 &  & 8.474 & 26.08 & 6.79 & 57.16 \\*
 & 12 &  & 11.864 & 24.95 & 6.42 & 58.63 \\*
 & 16 &  & 15.253 & 24.60 & 6.25 & 58.94 \\*
 & 32 &  & 28.810 & 24.39 & 6.14 & 59.21 \\
\midrule
 & 1 &  & 2.967 & 78.97 & 11.00 & 18.98 \\*
 & 2 &  & 4.238 & 46.80 & 8.62 & 35.51 \\*
 & 3 &  & 5.508 & 32.63 & 7.84 & 49.36 \\*
\raisebox{-.5\baselineskip}[0pt][0pt]{LiFT XL/2 R3} & 4 & \raisebox{-.5\baselineskip}[0pt][0pt]{50} & 6.779 & 26.70 & 7.28 & 57.76 \\*
 & 5 &  & 8.050 & 23.90 & 6.93 & 62.75 \\*
 & 8 &  & 11.863 & 21.96 & 6.40 & 66.05 \\*
 & 16 &  & 22.030 & 21.23 & 5.89 & 67.20 \\*
 & 32 &  & 42.365 & 21.23 & 5.83 & 67.27 \\
\midrule
 & 1 &  & 3.390 & 99.69 & 15.79 & 14.56 \\*
 & 2 &  & 5.085 & 49.72 & 10.07 & 32.10 \\*
 & 3 &  & 6.779 & 33.52 & 7.56 & 46.02 \\*
\raisebox{-.5\baselineskip}[0pt][0pt]{LiFT XL/2 R4} & 5 & \raisebox{-.5\baselineskip}[0pt][0pt]{50} & 10.168 & 26.01 & 5.62 & 55.15 \\*
 & 6 &  & 11.863 & 26.11 & 6.32 & 54.92 \\*
 & 8 &  & 15.252 & 23.66 & 5.41 & 58.32 \\*
 & 16 &  & 28.808 & 22.18 & 4.93 & 60.49 \\*
 & 32 &  & 55.919 & 21.78 & 4.76 & 61.32 \\
\midrule
 & 1 &  & 4.237 & 41.77 & 9.40 & 41.59 \\*
 & 2 &  & 6.779 & 23.67 & 7.33 & 64.82 \\*
 & 3 &  & 9.321 & 19.17 & 6.65 & 74.46 \\*
\raisebox{-.5\baselineskip}[0pt][0pt]{LiFT XL/2 R6} & 4 & \raisebox{-.5\baselineskip}[0pt][0pt]{50} & 11.862 & 18.08 & 6.22 & 76.97 \\*
 & 5 &  & 14.404 & 16.02 & 5.30 & 82.10 \\*
 & 8 &  & 22.029 & 13.11 & 4.55 & 91.00 \\*
 & 16 &  & 42.362 & 12.39 & 4.38 & 93.69 \\*
 & 32 &  & 83.029 & 12.20 & 4.33 & 94.54 \\
\midrule
 & 1 &  & 5.085 & 33.91 & 8.97 & 50.74 \\*
 & 2 &  & 8.473 & 19.93 & 6.95 & 74.82 \\*
 & 3 &  & 11.862 & 16.95 & 6.02 & 81.87 \\*
\raisebox{-.5\baselineskip}[0pt][0pt]{LiFT XL/2 R8} & 4 & \raisebox{-.5\baselineskip}[0pt][0pt]{50} & 15.251 & 11.36 & 4.33 & 100.08 \\*
 & 5 &  & 18.640 & 11.11 & 4.25 & 101.13 \\*
 & 8 &  & 28.806 & 9.93 & 4.02 & 107.47 \\*
 & 16 &  & 55.917 & 9.53 & 3.94 & 109.25 \\*
 & 32 &  & 110.138 & 9.50 & 3.93 & 109.29 \\
\midrule
 & 1 &  & 6.779 & 27.71 & 7.15 & 60.14 \\*
 & 2 &  & 11.862 & 17.07 & 6.41 & 82.66 \\*
 & 3 &  & 16.945 & 12.74 & 4.42 & 93.63 \\*
\raisebox{-.5\baselineskip}[0pt][0pt]{LiFT XL/2 R12} & 4 & \raisebox{-.5\baselineskip}[0pt][0pt]{50} & 22.028 & 10.69 & 4.13 & 102.06 \\*
 & 5 &  & 27.111 & 10.25 & 4.02 & 103.72 \\*
 & 8 &  & 42.361 & 9.55 & 3.96 & 106.79 \\*
 & 16 &  & 83.026 & 9.31 & 3.93 & 107.88 \\*
 & 32 &  & 164.356 & 9.71 & 3.89 & 106.20 \\
\end{longtable}
\endgroup

\FloatBarrier
\subsection{Depth-generalization matrices}
\label{app:depth_generalization}

\begin{figure}[htbp]
\centering
\includegraphics[width=\linewidth]{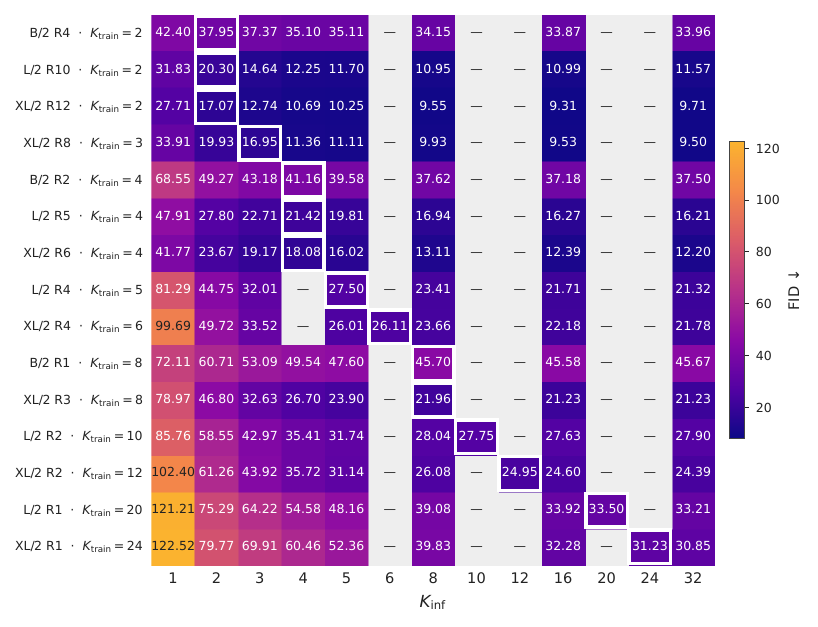}
\caption{FID of every checkpoint in the main depth sweeps across
inference depths, on a shared colour scale. The outlined cell in each row
marks training depth, and grey cells indicate unevaluated settings.}
\label{fig:depth_generalization_matrix}
\end{figure}

\Cref{fig:depth_generalization_matrix,tab:depth_generalization_fid}
report every measured point in the main depth sweeps,
with rows identifying checkpoints and columns specifying inference depth.
We express depth extrapolation through the ratio
\[
 r=\frac{K_{\mathrm{inf}}}{K_{\mathrm{train}}},
\]
so that evaluating a two-loop checkpoint at sixteen loops corresponds to
$r=8$. Because rows differ in backbone and core size as well as in
training depth, comparisons across rows also reflect the architecture.

\begin{table}[htbp]
\centering
\small
\setlength{\tabcolsep}{5pt}
\caption{\textbf{Depth extrapolation for all evaluated checkpoints.}
All settings use 50 integration steps and 50,000 generated images.
The best tested depth minimizes FID within each checkpoint's measured
sweep; $r=K_{\mathrm{inf}}/K_{\mathrm{train}}$ is evaluated at that depth.}
\label{tab:all_depth_extrapolation}
\begin{tabular}{lrrrrr}
\toprule
Model & $K_{\rm train}$ & FID at $K_{\rm train}$ & Best $K_{\rm inf}$ & Best FID & $r$ \\
\midrule
B/2 R4 & 2 & 37.95 & 16 & 33.87 & $8\times$ \\
L/2 R10 & 2 & 20.30 & 8 & 10.95 & $4\times$ \\
XL/2 R12 & 2 & 17.07 & 16 & 9.31 & $8\times$ \\
XL/2 R8 & 3 & 16.95 & 32 & 9.50 & $10.67\times$ \\
B/2 R2 & 4 & 41.16 & 16 & 37.18 & $4\times$ \\
L/2 R5 & 4 & 21.42 & 32 & 16.21 & $8\times$ \\
XL/2 R6 & 4 & 18.08 & 32 & 12.20 & $8\times$ \\
L/2 R4 & 5 & 27.50 & 32 & 21.32 & $6.4\times$ \\
XL/2 R4 & 6 & 26.11 & 32 & 21.78 & $5.33\times$ \\
B/2 R1 & 8 & 45.70 & 16 & 45.58 & $2\times$ \\
XL/2 R3 & 8 & 21.96 & 32 & 21.23 & $4\times$ \\
L/2 R2 & 10 & 27.75 & 16 & 27.63 & $1.6\times$ \\
XL/2 R2 & 12 & 24.95 & 32 & 24.39 & $2.67\times$ \\
L/2 R1 & 20 & 33.50 & 32 & 33.21 & $1.6\times$ \\
XL/2 R1 & 24 & 31.23 & 32 & 30.85 & $1.33\times$ \\
\bottomrule
\end{tabular}
\end{table}

\begin{table}[tb]
\centering\footnotesize
\setlength{\tabcolsep}{1.5pt}
\caption{FID of every checkpoint across inference depths, with the
depth ratio $r$ beneath each row. Bold values mark training depth, dashes
denote unevaluated settings, and $\dagger$ marks an increase in FID
relative to the preceding evaluated depth.}
\label{tab:depth_generalization_fid}
\begin{tabular}{llrrrrrrrrrrrrr}
\toprule
 & & \multicolumn{13}{c}{$K_{\mathrm{inf}}$} \\
Model & $K_{\rm train}$ & 1 & 2 & 3 & 4 & 5 & 6 & 8 & 10 & 12 & 16 & 20 & 24 & 32 \\
\midrule
B/2 R4 & 2 & 42.40 & \textbf{37.95} & 37.37 & 35.10 & 35.11$^{\dagger}$ & -- & 34.15 & -- & -- & 33.87 & -- & -- & 33.96$^{\dagger}$ \\
$r$ & -- & 0.50 & 1.00 & 1.50 & 2.00 & 2.50 & 3.00 & 4.00 & 5.00 & 6.00 & 8.00 & 10.00 & 12.00 & 16.00 \\
L/2 R10 & 2 & 31.83 & \textbf{20.30} & 14.64 & 12.25 & 11.70 & -- & 10.95 & -- & -- & 10.99$^{\dagger}$ & -- & -- & 11.57$^{\dagger}$ \\
$r$ & -- & 0.50 & 1.00 & 1.50 & 2.00 & 2.50 & 3.00 & 4.00 & 5.00 & 6.00 & 8.00 & 10.00 & 12.00 & 16.00 \\
XL/2 R12 & 2 & 27.71 & \textbf{17.07} & 12.74 & 10.69 & 10.25 & -- & 9.55 & -- & -- & 9.31 & -- & -- & 9.71$^{\dagger}$ \\
$r$ & -- & 0.50 & 1.00 & 1.50 & 2.00 & 2.50 & 3.00 & 4.00 & 5.00 & 6.00 & 8.00 & 10.00 & 12.00 & 16.00 \\
XL/2 R8 & 3 & 33.91 & 19.93 & \textbf{16.95} & 11.36 & 11.11 & -- & 9.93 & -- & -- & 9.53 & -- & -- & 9.50 \\
$r$ & -- & 0.33 & 0.67 & 1.00 & 1.33 & 1.67 & 2.00 & 2.67 & 3.33 & 4.00 & 5.33 & 6.67 & 8.00 & 10.67 \\
B/2 R2 & 4 & 68.55 & 49.27 & 43.18 & \textbf{41.16} & 39.58 & -- & 37.62 & -- & -- & 37.18 & -- & -- & 37.50$^{\dagger}$ \\
$r$ & -- & 0.25 & 0.50 & 0.75 & 1.00 & 1.25 & 1.50 & 2.00 & 2.50 & 3.00 & 4.00 & 5.00 & 6.00 & 8.00 \\
L/2 R5 & 4 & 47.91 & 27.80 & 22.71 & \textbf{21.42} & 19.81 & -- & 16.94 & -- & -- & 16.27 & -- & -- & 16.21 \\
$r$ & -- & 0.25 & 0.50 & 0.75 & 1.00 & 1.25 & 1.50 & 2.00 & 2.50 & 3.00 & 4.00 & 5.00 & 6.00 & 8.00 \\
XL/2 R6 & 4 & 41.77 & 23.67 & 19.17 & \textbf{18.08} & 16.02 & -- & 13.11 & -- & -- & 12.39 & -- & -- & 12.20 \\
$r$ & -- & 0.25 & 0.50 & 0.75 & 1.00 & 1.25 & 1.50 & 2.00 & 2.50 & 3.00 & 4.00 & 5.00 & 6.00 & 8.00 \\
L/2 R4 & 5 & 81.29 & 44.75 & 32.01 & -- & \textbf{27.50} & -- & 23.41 & -- & -- & 21.71 & -- & -- & 21.32 \\
$r$ & -- & 0.20 & 0.40 & 0.60 & 0.80 & 1.00 & 1.20 & 1.60 & 2.00 & 2.40 & 3.20 & 4.00 & 4.80 & 6.40 \\
XL/2 R4 & 6 & 99.69 & 49.72 & 33.52 & -- & 26.01 & \textbf{26.11}$^{\dagger}$ & 23.66 & -- & -- & 22.18 & -- & -- & 21.78 \\
$r$ & -- & 0.17 & 0.33 & 0.50 & 0.67 & 0.83 & 1.00 & 1.33 & 1.67 & 2.00 & 2.67 & 3.33 & 4.00 & 5.33 \\
B/2 R1 & 8 & 72.11 & 60.71 & 53.09 & 49.54 & 47.60 & -- & \textbf{45.70} & -- & -- & 45.58 & -- & -- & 45.67$^{\dagger}$ \\
$r$ & -- & 0.12 & 0.25 & 0.38 & 0.50 & 0.62 & 0.75 & 1.00 & 1.25 & 1.50 & 2.00 & 2.50 & 3.00 & 4.00 \\
XL/2 R3 & 8 & 78.97 & 46.80 & 32.63 & 26.70 & 23.90 & -- & \textbf{21.96} & -- & -- & 21.23 & -- & -- & 21.23 \\
$r$ & -- & 0.12 & 0.25 & 0.38 & 0.50 & 0.62 & 0.75 & 1.00 & 1.25 & 1.50 & 2.00 & 2.50 & 3.00 & 4.00 \\
L/2 R2 & 10 & 85.76 & 58.55 & 42.97 & 35.41 & 31.74 & -- & 28.04 & \textbf{27.75} & -- & 27.63 & -- & -- & 27.90$^{\dagger}$ \\
$r$ & -- & 0.10 & 0.20 & 0.30 & 0.40 & 0.50 & 0.60 & 0.80 & 1.00 & 1.20 & 1.60 & 2.00 & 2.40 & 3.20 \\
XL/2 R2 & 12 & 102.40 & 61.26 & 43.92 & 35.72 & 31.14 & -- & 26.08 & -- & \textbf{24.95} & 24.60 & -- & -- & 24.39 \\
$r$ & -- & 0.08 & 0.17 & 0.25 & 0.33 & 0.42 & 0.50 & 0.67 & 0.83 & 1.00 & 1.33 & 1.67 & 2.00 & 2.67 \\
L/2 R1 & 20 & 121.21 & 75.29 & 64.22 & 54.58 & 48.16 & -- & 39.08 & -- & -- & 33.92 & \textbf{33.50} & -- & 33.21 \\
$r$ & -- & 0.05 & 0.10 & 0.15 & 0.20 & 0.25 & 0.30 & 0.40 & 0.50 & 0.60 & 0.80 & 1.00 & 1.20 & 1.60 \\
XL/2 R1 & 24 & 122.52 & 79.77 & 69.91 & 60.46 & 52.36 & -- & 39.83 & -- & -- & 32.28 & -- & \textbf{31.23} & 30.85 \\
$r$ & -- & 0.04 & 0.08 & 0.12 & 0.17 & 0.21 & 0.25 & 0.33 & 0.42 & 0.50 & 0.67 & 0.83 & 1.00 & 1.33 \\
\bottomrule
\end{tabular}
\end{table}

\paragraph{Observed extrapolation.}
Several checkpoints improve substantially beyond training depth before
their gains diminish. L/2 R10, for example, reduces FID from 20.30 to
10.95 at $r=4$, and XL/2 R12 from 17.07 to 9.31 at $r=8$; both then
regress slightly at larger depths, whereas L/2 R5 and XL/2 R6 reach their
best tested FID at 32 loops. By contrast, the single-block cores, as well
as the two-block cores in L/2 and XL/2, gain little beyond training depth.

\FloatBarrier
\subsection{Joint allocation of inference computation}
\label{app:joint_budget}
\label{app:budget_envelopes}

Increasing recurrent depth and increasing the number of integration steps both add
sampling computation, but their effects need not be interchangeable. We
therefore evaluate LiFT L/2 R10 over the Cartesian grid
\[
 T\in\{10,25,50,100\},\qquad
 K_{\mathrm{inf}}\in\{1,2,4,8,16,32\}.
\]
Because all settings use the same frozen 500,000-update checkpoint, trained at
$K=2$ with four prelude blocks, ten core blocks, and auxiliary
prelude-loss weight $\lambda=0.01$, the sweep holds trainable parameters,
training tokens, and training compute fixed. Every setting follows the
evaluation protocol of \cref{app:evaluation_protocol}, so initial noise and
class labels are paired across settings.

\Cref{sec:compute_allocation} draws its main allocation results from
this grid: four loops at 25 integration steps improve on the two-loop
training setting at lower cost, whereas sixteen loops at ten steps
are both worse and more expensive than that setting.

\paragraph{Diminishing returns.}
The gains from both recurrence and integration steps diminish at larger
budgets. At eight loops, increasing the number of steps from 25 to 50 reduces
FID from 11.39 to 10.95, whereas doubling the steps again to 100 leaves
FID nearly unchanged at 10.96. Eight loops is the best tested depth at all
three of these step counts, while extending the rollout from eight to
32 loops worsens FID at all four budgets.

\paragraph{Metric dependence.}
Even at a fixed number of steps, the preferred depth depends on the quality
metric. At 25 integration steps, for example, increasing depth from eight
to 32 improves sFID from 4.36 to 3.96, even as FID worsens from 11.39 to
12.14 and IS decreases from 97.79 to 94.34.

\begin{figure}[htbp]
\centering
\includegraphics[width=.9\linewidth]{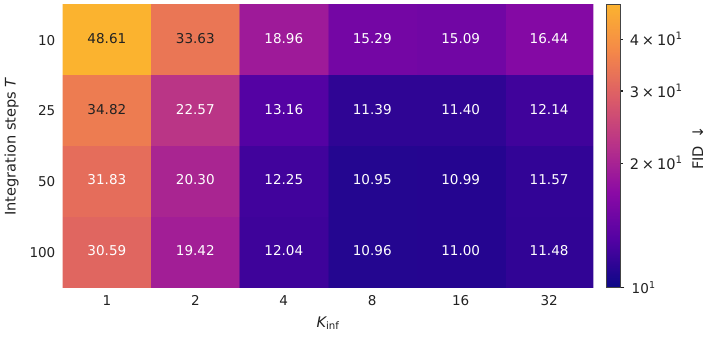}
\caption{FID across integration steps and recurrent depth. Colour encodes FID on a logarithmic scale.}
\label{fig:joint_budget_heatmap}
\end{figure}

\begingroup
\footnotesize
\begin{longtable}{rrrrrr}
\caption{Complete joint-sweep operating points for LiFT L/2 R10, including
additional depths 3 and 5 at 50 integration steps. All metrics in a row use the same images.}
\label{tab:joint_budget_points}\\
\toprule $T$ & $K_{\rm inf}$ & TFLOPs/image & FID & sFID & IS \\
\midrule\endfirsthead
\multicolumn{6}{l}{\tablename~\thetable{} (continued)}\\
\toprule $T$ & $K_{\rm inf}$ & TFLOPs/image & FID & sFID & IS \\
\midrule\endhead
\bottomrule\endfoot
\multirow{6}{*}{10} & 1 & 0.942 & 48.613 & 10.751 & 41.770 \\*
 & 2 & 1.614 & 33.633 & 11.497 & 58.830 \\*
 & 4 & 2.959 & 18.955 & 6.107 & 81.803 \\*
 & 8 & 5.648 & 15.289 & 5.428 & 90.433 \\*
 & 16 & 11.027 & 15.094 & 5.481 & 90.670 \\*
 & 32 & 21.785 & 16.435 & 5.606 & 86.694 \\
\midrule
\multirow{6}{*}{25} & 1 & 2.354 & 34.819 & 6.898 & 49.804 \\*
 & 2 & 4.035 & 22.570 & 7.270 & 69.810 \\*
 & 4 & 7.397 & 13.165 & 4.041 & 91.327 \\*
 & 8 & 14.120 & 11.390 & 4.360 & 97.795 \\*
 & 16 & 27.568 & 11.402 & 4.308 & 97.103 \\*
 & 32 & 54.463 & 12.142 & 3.963 & 94.345 \\
\midrule
\multirow{8}{*}{50} & 1 & 4.708 & 31.828 & 6.223 & 51.216 \\*
 & 2 & 8.070 & 20.304 & 6.345 & 71.805 \\*
 & 3 & 11.431 & 14.642 & 4.217 & 84.120 \\*
 & 4 & 14.793 & 12.255 & 4.123 & 92.013 \\*
 & 5 & 18.155 & 11.702 & 4.326 & 93.526 \\*
 & 8 & 28.241 & 10.952 & 4.744 & 97.169 \\*
 & 16 & 55.136 & 10.994 & 4.615 & 96.865 \\*
 & 32 & 108.926 & 11.567 & 4.074 & 93.746 \\
\midrule
\multirow{6}{*}{100} & 1 & 9.415 & 30.586 & 5.933 & 51.696 \\*
 & 2 & 16.139 & 19.422 & 5.926 & 72.730 \\*
 & 4 & 29.587 & 12.043 & 4.372 & 91.961 \\*
 & 8 & 56.482 & 10.961 & 5.180 & 95.980 \\*
 & 16 & 110.271 & 10.999 & 5.002 & 95.332 \\*
 & 32 & 217.851 & 11.484 & 4.330 & 93.193 \\
\end{longtable}
\endgroup

\FloatBarrier
\paragraph{Measured operating-point envelopes.}

To compare these allocations within an inference budget $B$, we select
the measured LiFT operating point with the lowest FID:
\[
 (n^*,k^*)\in\underset{(n,k)\in\mathcal G:\,
 n C_{\mathrm{forward}}(k)\le B}{\operatorname{argmin}}
 \operatorname{FID}(n,k),
\]
where $\mathcal G$ contains all measured operating points from the joint
sweep, together with the additional depths $K_{\mathrm{inf}}=3,5$
evaluated with 50 integration steps. Selecting within this set gives an
observed FID, step count, and recurrent depth for each budget, without interpolating between
settings. We apply the same budget constraint to dense B/2 and L/2 over
$T\in\{1,10,25,50,100,250\}$, using the analytic FLOP
accounting of the main figures and excluding VAE decoding.

LiFT L/2 R10 and dense L/2 also have comparable training computation,
with an estimated
LiFT overhead of $0.019\%$ (\cref{app:trajectory_compute}), whereas
dense B/2 has a smaller training budget.

\Cref{fig:joint_budget_frontier} shows these envelopes in the main text,
while \cref{tab:budget_envelopes} records the setting selected at each
measured budget.

\begingroup
\footnotesize
\setlength{\tabcolsep}{4pt}
\begin{longtable}{rrrrlr}
\caption{FID-minimizing measured settings within each inference budget.
The dense column selects among both evaluated dense backbones. Budgets
are rounded for display; nearby distinct costs may share a printed value.}
\label{tab:budget_envelopes}\\
\toprule Budget (TFLOPs) & LiFT $T$ & LiFT $K$ & LiFT FID & Dense setting & Dense FID \\
\midrule\endfirsthead
\multicolumn{6}{l}{\tablename~\thetable{} (continued)}\\
\toprule Budget (TFLOPs) & LiFT $T$ & LiFT $K$ & LiFT FID & Dense setting & Dense FID \\
\midrule\endhead
\bottomrule\endfoot
0.046 &  &  & -- & Dense B/2, $T{=}1$ & 321.169 \\*
0.161 & -- & -- & -- & Dense L/2, $T{=}1$ & 321.150 \\*
0.460 &  &  & -- & Dense B/2, $T{=}10$ & 44.486 \\
\midrule
0.942 &  &  & 48.613 & Dense B/2, $T{=}10$ & 44.486 \\*
1.150 &  & 1 & 48.613 & Dense B/2, $T{=}25$ & 32.501 \\*
1.614 &  &  & 48.613 & Dense L/2, $T{=}10$ & 29.591 \\*
\cmidrule(lr){1-1}\cmidrule(lr){3-6}
1.614 &  &  & 33.633 & Dense L/2, $T{=}10$ & 29.591 \\*
2.300 &  & 2 & 33.633 & Dense L/2, $T{=}10$ & 29.591 \\*
2.354 & 10 &  & 33.633 & Dense L/2, $T{=}10$ & 29.591 \\*
\cmidrule(lr){1-1}\cmidrule(lr){3-6}
2.959 &  &  & 18.955 & Dense L/2, $T{=}10$ & 29.591 \\*
4.035 &  &  & 18.955 & Dense L/2, $T{=}25$ & 19.047 \\*
4.035 &  & 4 & 18.955 & Dense L/2, $T{=}25$ & 19.047 \\*
4.600 &  &  & 18.955 & Dense L/2, $T{=}25$ & 19.047 \\*
4.708 &  &  & 18.955 & Dense L/2, $T{=}25$ & 19.047 \\*
\cmidrule(lr){1-1}\cmidrule(lr){3-6}
5.648 &  & 8 & 15.289 & Dense L/2, $T{=}25$ & 19.047 \\
\midrule
7.397 &  &  & 13.165 & Dense L/2, $T{=}25$ & 19.047 \\*
8.069 &  &  & 13.165 & Dense L/2, $T{=}50$ & 16.940 \\*
8.070 &  &  & 13.165 & Dense L/2, $T{=}50$ & 16.940 \\*
9.415 &  & 4 & 13.165 & Dense L/2, $T{=}50$ & 16.940 \\*
11.027 &  &  & 13.165 & Dense L/2, $T{=}50$ & 16.940 \\*
11.431 &  &  & 13.165 & Dense L/2, $T{=}50$ & 16.940 \\*
11.501 & 25 &  & 13.165 & Dense L/2, $T{=}50$ & 16.940 \\*
\cmidrule(lr){1-1}\cmidrule(lr){3-6}
14.120 &  &  & 11.390 & Dense L/2, $T{=}50$ & 16.940 \\*
14.793 &  &  & 11.390 & Dense L/2, $T{=}50$ & 16.940 \\*
16.139 &  &  & 11.390 & Dense L/2, $T{=}100$ & 16.173 \\*
16.139 &  & 8 & 11.390 & Dense L/2, $T{=}100$ & 16.173 \\*
18.155 &  &  & 11.390 & Dense L/2, $T{=}100$ & 16.173 \\*
21.785 &  &  & 11.390 & Dense L/2, $T{=}100$ & 16.173 \\*
27.568 &  &  & 11.390 & Dense L/2, $T{=}100$ & 16.173 \\
\midrule
28.241 &  &  & 10.952 & Dense L/2, $T{=}100$ & 16.173 \\*
29.587 &  &  & 10.952 & Dense L/2, $T{=}100$ & 16.173 \\*
40.347 &  &  & 10.952 & Dense L/2, $T{=}250$ & 15.743 \\*
54.463 &  &  & 10.952 & Dense L/2, $T{=}250$ & 15.743 \\*
55.136 & 50 & 8 & 10.952 & Dense L/2, $T{=}250$ & 15.743 \\*
56.482 &  &  & 10.952 & Dense L/2, $T{=}250$ & 15.743 \\*
108.926 &  &  & 10.952 & Dense L/2, $T{=}250$ & 15.743 \\*
110.271 &  &  & 10.952 & Dense L/2, $T{=}250$ & 15.743 \\*
217.851 &  &  & 10.952 & Dense L/2, $T{=}250$ & 15.743 \\
\end{longtable}
\endgroup

\FloatBarrier

\FloatBarrier
\section{Ablations and diagnostics}
\label{app:ablations}

This appendix first ablates two training choices, the depth coordinates
and the auxiliary prelude loss, and then examines how a trained model
responds to changes in its inference grid. The training ablations share
one setup: LiFT B/2 with four prelude blocks, a two-block shared core,
four training loops, and no transformer coda. Each setting is trained
once for 100,000 updates at a global batch size of 256, otherwise
following the pipeline and optimization settings of
\cref{app:experimental-details}, and generation follows the evaluation
protocol of \cref{app:evaluation_protocol}.
\Cref{app:prediction_trajectories,app:grid_consistency} then turn to a
frozen L/2 checkpoint and ask how its intermediate predictions and final
velocities vary with the inference grid.

\subsection{Random versus fixed depth coordinates}
\label{app:depth_coordinates}

This comparison isolates the effect of sampling the depth coordinates at
random. The fixed-grid variant replaces the random interior coordinates
with the uniform grid $s_k=k/K_{\mathrm{train}}$ and uses it both to
condition the core and to construct the reference targets, while the
random-grid variant keeps its sampled coordinates for both. Everything
else is shared: the two runs use the same architecture, depth embeddings,
and trajectory supervision, and both are evaluated on the uniform grid
$s_k=k/K_{\mathrm{inf}}$. To keep the comparison free of auxiliary prelude
regression, both variants set $\lambda=0$, whereas the main experiments
use $\lambda=0.01$.

\begin{figure}[htbp]
\centering
\includegraphics[width=\linewidth]{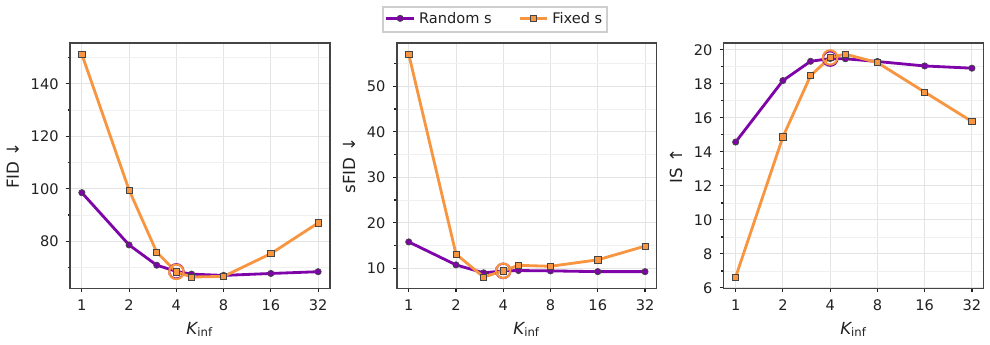}
\caption{\textbf{Training-grid randomization improves depth robustness.}
Both models use trajectory supervision with $\lambda=0$ and are evaluated
on uniform grids ending at $s=1$. Each point uses the same 50,000 initial
noise samples and class labels, EMA weights, 50 Euler steps, and no
classifier-free guidance. Hollow rings mark the training depth, $K=4$.}
\label{fig:depth_conditioning}
\end{figure}

\paragraph{Random coordinates improve robustness away from the training depth.}
At four loops, the two checkpoints attain similar FID: 68.51 with random
coordinates and 68.31 with fixed coordinates. The fixed-grid model reaches
its best tested FID of 66.35 at five loops, slightly below the random-grid
model's best of 66.97 at eight loops. The difference between the two
models becomes clearer at larger depths, where fixed-grid FID rises to
75.26 at 16 loops and 86.95 at 32, while the random-grid model stays close
to its best, at 67.71 and 68.38, respectively
(\cref{fig:depth_conditioning}). Random coordinates also
improve FID at depths below the training budget, particularly at one and
two loops.

sFID and IS show the same pattern at large depth: at 32 loops, random
coordinates give sFID 9.26 and IS 18.92, compared with 14.87 and 15.78 for
fixed coordinates. \Cref{tab:depth_conditioning} lists every evaluated
setting.

\begin{table}[htbp]
\centering
\small
\setlength{\tabcolsep}{5pt}
\caption{\textbf{Random versus fixed training coordinates.} Complete metrics
for B/2 R2 with $K_{\mathrm{train}}=4$, $\lambda=0$, and 100,000 updates.
All settings use 50,000 images and 50 integration steps.}
\label{tab:depth_conditioning}
\begin{tabular}{rrrrrrr}
\toprule
& \multicolumn{3}{c}{Random $s$} & \multicolumn{3}{c}{Fixed $s$} \\
\cmidrule(lr){2-4}\cmidrule(lr){5-7}
$K_{\mathrm{inf}}$ & FID $\downarrow$ & sFID $\downarrow$ & IS $\uparrow$ & FID $\downarrow$ & sFID $\downarrow$ & IS $\uparrow$ \\
\midrule
1 & 98.48 & 15.80 & 14.58 & 151.23 & 57.03 & 6.64 \\
2 & 78.57 & 10.75 & 18.19 & 99.47 & 13.12 & 14.88 \\
3 & 70.87 & 9.00 & 19.32 & 75.81 & 8.07 & 18.48 \\
4 & 68.51 & 9.43 & 19.49 & 68.31 & 9.45 & 19.55 \\
5 & 67.46 & 9.51 & 19.47 & 66.35 & 10.65 & 19.74 \\
8 & 66.97 & 9.43 & 19.31 & 66.60 & 10.44 & 19.25 \\
16 & 67.71 & 9.26 & 19.05 & 75.26 & 11.86 & 17.51 \\
32 & 68.38 & 9.26 & 18.92 & 86.95 & 14.87 & 15.78 \\
\bottomrule
\end{tabular}
\end{table}

\subsection{Prelude-loss / \texorpdfstring{$\lambda$}{lambda} ablation}
\label{app:prelude_loss}

The prelude-loss sweep keeps trajectory supervision and random depth
coordinates and varies only $\lambda\in\{0,0.01,0.1,1\}$; its $\lambda=0$
run is the random-grid run of \cref{app:depth_coordinates}. As in
\cref{eq:lift_training_loss}, the auxiliary term is added once, outside
the average over recurrent predictions. Its role is to keep the initial
prediction, and hence the anchor, at the scale of the target, so we track
the anchor's magnitude and error throughout training.

\begin{figure}[htbp]
\centering
\includegraphics[width=\linewidth]{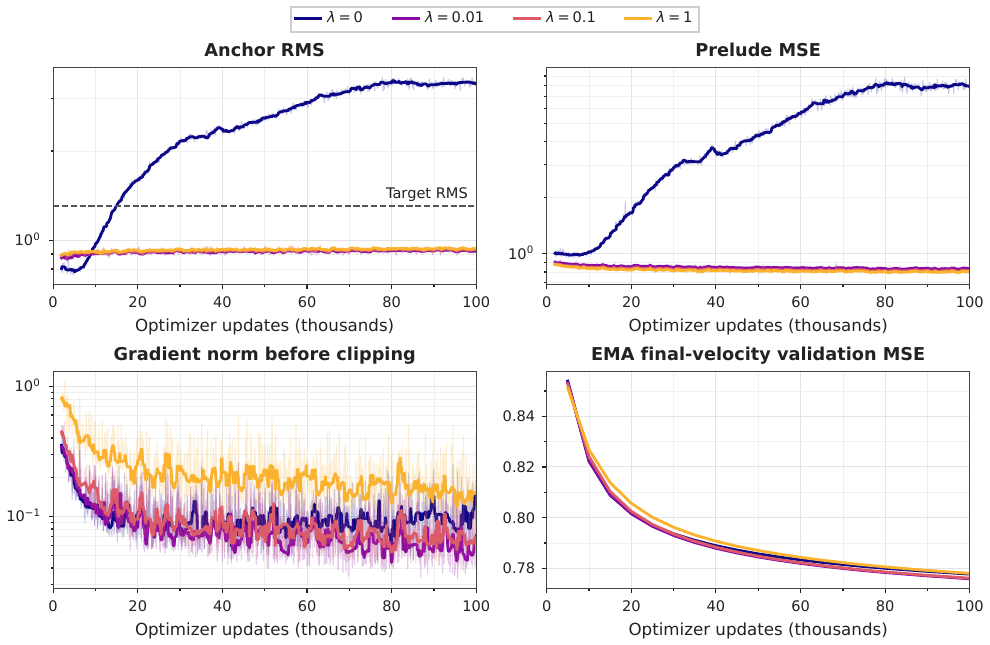}
\caption{\textbf{Prelude supervision controls the scale of the initial
prediction.} LiFT B/2 R2, trained with $K=4$ for 100,000 updates, varying
only the auxiliary weight $\lambda$. Training panels show logged values
faintly and an 11-point moving median prominently, omitting the first
2,000 updates; validation curves show unsmoothed full-validation EMA
velocity MSE. The dashed line marks the approximate target RMS.}
\label{fig:lambda_training_curves}
\end{figure}

\paragraph{A small auxiliary weight controls anchor drift.}
Without prelude supervision, the anchor RMS rises throughout training,
reaching 3.39 over the final 10,000 updates against a target RMS of 1.30,
and its median MSE to the flow-matching target grows to 7.96. All three
nonzero weights prevent this drift, keeping the anchor RMS near 0.93 and
its MSE below 0.84 (\cref{fig:lambda_training_curves} and
\cref{tab:lambda_training_diagnostics}), so even $\lambda=0.01$ suffices.

\begin{table}[htbp]
\centering
\small
\setlength{\tabcolsep}{5pt}
\caption{\textbf{Prelude-loss training diagnostics.} Anchor RMS, prelude
MSE, and gradient norm are medians of logged values over updates
90,000--100,000. Validation MSE uses EMA weights at update 100,000.}
\label{tab:lambda_training_diagnostics}
\begin{tabular}{rrrrr}
\toprule
$\lambda$ & Anchor RMS & Prelude MSE & Gradient norm & Validation MSE \\
\midrule
0 & 3.3875 & 7.9626 & 0.0978 & 0.777666 \\
0.01 & 0.9238 & 0.8316 & 0.0568 & 0.775813 \\
0.1 & 0.9311 & 0.8100 & 0.0661 & 0.776069 \\
1 & 0.9380 & 0.7992 & 0.1525 & 0.777948 \\
\bottomrule
\end{tabular}
\end{table}

\begin{figure}[htbp]
\centering
\includegraphics[width=\linewidth]{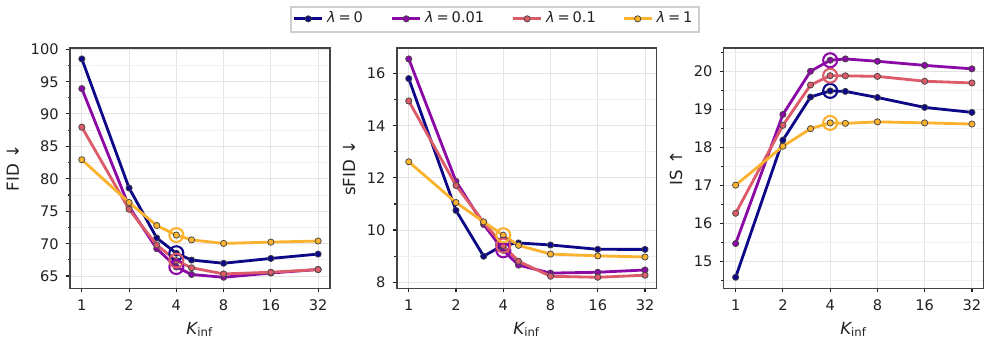}
\caption{\textbf{Generation quality across prelude-loss weights.}
FID, sFID, and IS for the same four 100,000-update checkpoints used in
\cref{fig:lambda_training_curves}. Each point uses 50,000 generated
images with 50 integration steps, with common initial noise and class labels. Hollow rings
mark the training depth, $K=4$.}
\label{fig:lambda_generation}
\end{figure}

\paragraph{Larger weights bring no further benefit.}
Stronger prelude regression reduces the prelude's own error further but
does not help downstream. The final validation MSE is lowest at
$\lambda=0.01$, if only by about 0.24\% relative to the zero-weight run
(\cref{tab:lambda_training_diagnostics}), and FID is lowest there as
well: 64.81 at $K_{\mathrm{inf}}=8$, compared with 66.97 without prelude
supervision, 65.32 with $\lambda=0.1$, and 70.03 with $\lambda=1$
(\cref{fig:lambda_generation}). We therefore use $\lambda=0.01$
throughout, and \cref{tab:lambda_generation} lists every evaluated
depth.

\begin{table}[htbp]
\centering
\scriptsize
\setlength{\tabcolsep}{4pt}
\caption{\textbf{Complete prelude-loss sweep.} All runs use the same B/2 R2
architecture, $K_{\mathrm{train}}=4$, and 100,000 updates. Each setting uses
50,000 images, EMA weights, 50 Euler steps, and no classifier-free guidance.}
\label{tab:lambda_generation}
\begin{tabular}{rlrrrrrrrr}
\toprule
$\lambda$ & Metric & $K=1$ & 2 & 3 & 4 & 5 & 8 & 16 & 32 \\
\midrule
0 & FID & 98.48 & 78.57 & 70.87 & 68.51 & 67.46 & 66.97 & 67.71 & 68.38 \\
 & sFID & 15.80 & 10.75 & 9.00 & 9.43 & 9.51 & 9.43 & 9.26 & 9.26 \\
 & IS & 14.58 & 18.19 & 19.32 & 19.49 & 19.47 & 19.31 & 19.05 & 18.92 \\
\midrule
0.01 & FID & 93.90 & 75.69 & 69.22 & 66.39 & 65.24 & 64.81 & 65.46 & 66.01 \\
 & sFID & 16.55 & 11.86 & 10.21 & 9.23 & 8.66 & 8.35 & 8.39 & 8.47 \\
 & IS & 15.47 & 18.87 & 20.00 & 20.29 & 20.33 & 20.27 & 20.16 & 20.07 \\
\midrule
0.1 & FID & 87.96 & 75.29 & 69.79 & 67.38 & 66.27 & 65.32 & 65.59 & 65.99 \\
 & sFID & 14.94 & 11.71 & 10.30 & 9.41 & 8.81 & 8.23 & 8.19 & 8.27 \\
 & IS & 16.26 & 18.58 & 19.64 & 19.89 & 19.88 & 19.87 & 19.74 & 19.69 \\
\midrule
1 & FID & 82.94 & 76.34 & 72.79 & 71.31 & 70.57 & 70.03 & 70.23 & 70.40 \\
 & sFID & 12.61 & 11.05 & 10.31 & 9.80 & 9.41 & 9.08 & 9.01 & 8.97 \\
 & IS & 17.00 & 18.03 & 18.49 & 18.64 & 18.63 & 18.67 & 18.64 & 18.61 \\
\bottomrule
\end{tabular}
\end{table}

\subsection{Prediction trajectories}
\label{app:prediction_trajectories}

Whereas the generation experiments measure how additional recurrence
affects sample quality, this analysis looks inside a single evaluation of
the velocity field. We first compare the intermediate readouts with the
supervised reference path toward the sampled target, and then with the
chord joining the model's own initial and final estimates.

\paragraph{Paired evaluation protocol.}
We use the EMA weights of the frozen LiFT L/2 R10 checkpoint from
\cref{app:joint_budget}, trained with $K_{\mathrm{train}}=2$. From
the ImageNet validation set, we select 512 unique examples, excluding
exact source duplicates of training images, and evaluate each at
$t\in\{0.05,0.25,0.5,0.75,0.95\}$. Every depth and grid receives the same
image, noise, class label, and noisy state $x_t=(1-t)x_0+t x_1$.
Uniform inference grids use $K_{\mathrm{inf}}\in\{1,2,3,4,5,8,16,32\}$;
the trajectory plots show five representative depths from two to 32.
We average measurements over generative times within each image before
summarizing across images, with approximate 95\% intervals given by
1.96 standard errors.

\begin{figure}[htbp]
    \centering
    \includegraphics[width=\linewidth]{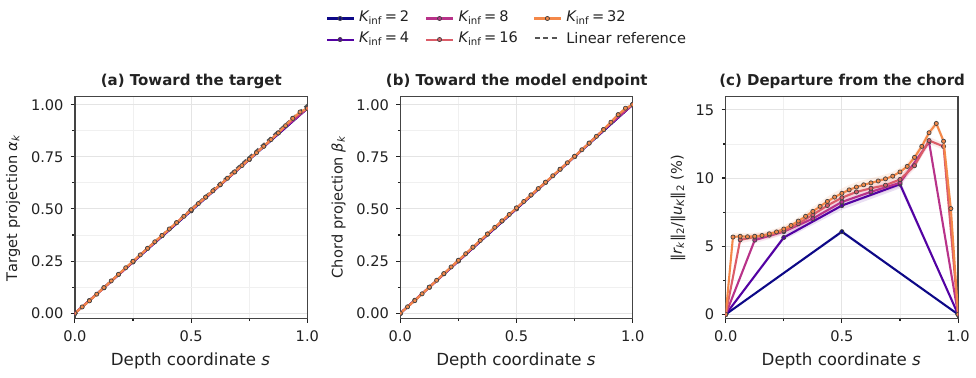}
    \caption{Prediction trajectories of LiFT L/2 R10, trained with two loops.
    (a) Projected progress toward the sampled target follows the assigned
    depth coordinate. (b) Projection onto each rollout's own endpoint chord
    is close to linear, and (c) the orthogonal residual remains similar
    across depths. Curves average five generative times within each of
    512 images; shaded bands show approximate 95\% intervals across images.
    Chord endpoints agree by construction.}
    \label{fig:prediction_trajectories}
\end{figure}

\paragraph{Progress toward the supervised target.}
For each input, let $b=u_0$ and $\Delta=u^\star-b$. We decompose the
displacement of readout $k$ into progress along the target direction and
an orthogonal residual:
\begin{equation}
    \alpha_k=\frac{\langle u_k-b,\Delta\rangle}{\|\Delta\|_2^2},
    \qquad
    e_k=(u_k-b)-\alpha_k\Delta.
    \label{eq:diagnostic_target_projection}
\end{equation}
The reference path has $\alpha_k=s_k$ and $e_k=0$. At uniform depth four,
the measured mean projections are $0.246$, $0.493$, $0.737$, and $0.982$
at $s=0.25,0.5,0.75,1$, respectively. At the same coordinates, depth
32 gives $0.249$, $0.495$, $0.738$, and $0.982$, showing that this
depth-indexed progression persists with substantially more recurrent
applications than were used in training
(\cref{fig:prediction_trajectories}a).

\paragraph{Geometry between the model's own predictions.}
To examine the realized path without reference to the sampled target, we
replace $\Delta$ by the chord $q=u_K-b$ of each rollout and define
\begin{equation}
    \beta_k=\frac{\langle u_k-b,q\rangle}{\|q\|_2^2},
    \qquad
    r_k=(u_k-b)-\beta_k q.
    \label{eq:diagnostic_model_chord}
\end{equation}
The projection $\beta_k$ tracks $s_k$ closely across the tested depths
(\cref{fig:prediction_trajectories}b): the mean absolute discrepancy
over interior readouts is $0.0006$ at $K=2$ and $0.0051$ at $K=32$.
The orthogonal residual is similar across depths as well: it is already
$6.07\%$ of the final prediction norm at the training depth $K=2$ and
reaches $8.68\%$ at $K=32$ (\cref{fig:prediction_trajectories}c), so
running sixteen times as many loops changes the shape of the trajectory
only slightly. \Cref{tab:prediction_chord} reports absolute and
normalized residuals over the interior readouts.

\begin{table}[htbp]
\centering\small
\setlength{\tabcolsep}{5pt}
\renewcommand{\arraystretch}{1.1}
\caption{Intermediate deviations from each rollout's own endpoint chord. We average over interior readouts, then generative times and images; the endpoints are excluded because they lie on the chord by construction. The last two columns use different normalization scales.}
\label{tab:prediction_chord}
\begin{tabular}{@{}rrrrr@{}}
\toprule
$K_{\mathrm{inf}}$ & $|\beta_k-s_k|$ & $\|r_k\|_2/\sqrt d$ & $100\|r_k\|_2/\|u_K-b\|_2$ & $100\|r_k\|_2/\|u_K\|_2$\\
\midrule
2 & 0.0006 & 0.0556 & 0.56 & 6.07\\
4 & 0.0030 & 0.0710 & 0.64 & 7.72\\
8 & 0.0040 & 0.0764 & 0.68 & 8.29\\
16 & 0.0049 & 0.0782 & 0.70 & 8.48\\
32 & 0.0051 & 0.0800 & 0.72 & 8.68\\
\bottomrule
\end{tabular}
\end{table}

\begin{bluebox}{}
\textbf{A learned progression beyond training depth.}
For this two-loop checkpoint, projections of intermediate predictions
onto their own endpoint chord remain almost linear in the depth
coordinate through 32 inference loops.
\end{bluebox}

\subsection{Computational-grid consistency}
\label{app:grid_consistency}

Although the preceding analysis shows a regular progression within each
rollout, different intermediate coordinates could still lead to different
endpoints. We therefore perturb the computational grid while holding the
generative state fixed, testing whether the final velocity remains
consistent as the route through computational depth changes.

\paragraph{Protocol.}
We retain the checkpoint, 512 paired examples, and five generative times
from \cref{app:prediction_trajectories}. At inference depth four,
we compare the uniform grid $(0.25,0.5,0.75,1)$ with front-loaded
$(0.1,0.2,0.4,1)$, back-loaded $(0.6,0.8,0.9,1)$, and three independently
seeded random-grid variants. We construct each random grid by sorting
three uniform draws and appending the endpoint one; the same per-image
grid is used at all five generative times. Uniform grids at depths
$1,2,3,5,8,16,32$ additionally test
sensitivity to the number of intermediate predictions. The initial depth
coordinate is zero in every case.

\paragraph{Endpoint agreement at fixed depth.}
For grid $g$, let $u^{(g)}$ denote its final velocity and $u^{(4)}$ the
prediction on the uniform four-step grid. We measure
\begin{equation}
    \delta_g=\frac{\|u^{(g)}-u^{(4)}\|_2}{\|u^{(4)}\|_2},
    \qquad
    c_g=\frac{\langle u^{(g)},u^{(4)}\rangle}
    {\|u^{(g)}\|_2\|u^{(4)}\|_2}.
    \label{eq:grid_endpoint_agreement}
\end{equation}
The front-loaded and back-loaded grids alter the endpoint by a mean
$1.89\%$ and $1.82\%$, respectively, with mean cosine similarity above
$0.9997$. The three random-grid variants give similar mean differences
of $1.72$--$1.85\%$, but have heavier tails: the 95th percentiles of
per-image time-averaged
differences are $4.26$--$4.61\%$, compared with $2.58$--$2.66\%$ for the two fixed
nonuniform grids (\cref{tab:grid_consistency}).

\begin{table}[htbp]
\centering\small
\setlength{\tabcolsep}{5pt}
\renewcommand{\arraystretch}{1.1}
\caption{Paired endpoint agreement with the uniform $K=4$ prediction. Differences are percentages of the reference prediction norm; means have approximate 95\% intervals across 512 images. The image-mean p95 averages the five times within each image before taking the percentile, whereas the worst-time p95 first takes their maximum. All nonuniform grids have $K=4$. MSE is measured against $u^\star$.}
\label{tab:grid_consistency}
\begin{tabular}{@{}lrrrrr@{}}
\toprule
Grid & Mean (\%) & Image-mean p95 & Worst-time p95 & Cosine & Target MSE\\
\midrule
Uniform $K=1$ & $11.01 \pm 0.20$ & 14.88 & 26.86 & 0.99198 & 0.7765\\
Uniform $K=2$ & $5.12 \pm 0.12$ & 7.63 & 16.59 & 0.99805 & 0.7669\\
Uniform $K=3$ & $2.61 \pm 0.07$ & 4.08 & 8.92 & 0.99948 & 0.7668\\
Uniform $K=4$ & $0.00 \pm 0.00$ & 0.00 & 0.00 & 1.00000 & 0.7670\\
Uniform $K=5$ & $1.70 \pm 0.05$ & 2.80 & 6.64 & 0.99977 & 0.7675\\
Uniform $K=8$ & $2.65 \pm 0.10$ & 4.88 & 13.98 & 0.99927 & 0.7684\\
Uniform $K=16$ & $3.66 \pm 0.14$ & 6.85 & 18.94 & 0.99872 & 0.7689\\
Uniform $K=32$ & $4.24 \pm 0.13$ & 7.03 & 18.72 & 0.99850 & 0.7686\\
\midrule
Front-loaded & $1.89 \pm 0.04$ & 2.66 & 4.72 & 0.99977 & 0.7667\\
Back-loaded & $1.82 \pm 0.04$ & 2.58 & 4.84 & 0.99977 & 0.7673\\
Random 1 & $1.85 \pm 0.11$ & 4.61 & 7.43 & 0.99968 & 0.7671\\
Random 2 & $1.82 \pm 0.11$ & 4.51 & 6.95 & 0.99970 & 0.7671\\
Random 3 & $1.72 \pm 0.10$ & 4.26 & 7.01 & 0.99972 & 0.7671\\
\bottomrule
\end{tabular}
\end{table}

\begin{figure}[htbp]
    \centering
    \includegraphics[width=\linewidth]{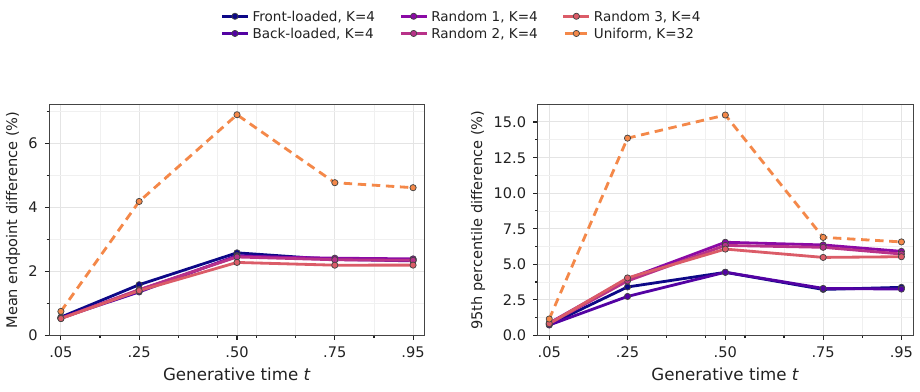}
    \caption{Sensitivity of the final velocity to the computational grid,
    relative to uniform $K=4$ at the identical noisy input. The left panel
    shows the mean relative distance and the right shows its 95th percentile across
    512 images at each generative time. Solid curves change coordinates
    while holding $K=4$ fixed; the dashed curve also increases the number
    of recurrent applications to 32.}
    \label{fig:grid_endpoint_times}
\end{figure}

\paragraph{Changing the number of recurrent applications.}
Relative to uniform $K=4$, the two-step endpoint differs by $5.12\%$ on
average, and the 32-step endpoint by $4.24\%$ (mean cosine $0.9985$),
even at 16 times the training depth. These averages conceal variation
with generative time (\cref{fig:grid_endpoint_times}): at $K=32$
and $t=0.5$, the mean difference is $6.90\%$ and its 95th percentile
$15.48\%$. Taking each image's largest difference across the five times
at this depth gives a
95th percentile of $18.72\%$. Target MSE, by contrast, barely moves,
staying within $0.7667$--$0.7689$ for all grids with $K\geq2$.

\paragraph{Numerical precision.}
We recompute the uniform four-step endpoint in FP32 for the same first 16
examples at each time and compare it with the BF16 prediction, yielding
pooled RMSE $\epsilon$ between $0.0065$ and $0.0150$
(\cref{tab:grid_precision}). On these identical examples, fixed-depth
grid differences are comparable to this numerical scale near $t=0.05$
but generally exceed it at intermediate times. Among the listed grids,
the largest ratio of grid-difference RMSE to $\epsilon$ is $7.5$, for
uniform depth 32 at $t=0.5$, so precision explains only the smallest of
the observed differences.

\begin{table}[htbp]
\centering\small
\setlength{\tabcolsep}{5pt}
\renewcommand{\arraystretch}{1.1}
\caption{Numerical precision check on the same 16 examples at each time. The second column gives pooled BF16--FP32 endpoint RMSE for uniform $K=4$. The remaining columns divide each grid's pooled RMSE relative to that BF16 reference by the numerical baseline, using the identical examples.}
\label{tab:grid_precision}
\begin{tabular}{@{}rrrrrr@{}}
\toprule
$t$ & BF16--FP32 RMSE & Front/$\epsilon$ & Back/$\epsilon$ & Random 1/$\epsilon$ & $K=32$/$\epsilon$\\
\midrule
0.05 & 0.00663 & 1.2 & 0.9 & 0.8 & 1.2\\
0.25 & 0.00654 & 3.5 & 4.3 & 2.4 & 5.9\\
0.50 & 0.00717 & 3.5 & 3.0 & 2.7 & 7.5\\
0.75 & 0.00937 & 2.4 & 2.5 & 1.9 & 4.9\\
0.95 & 0.01500 & 1.5 & 1.4 & 1.1 & 2.8\\
\bottomrule
\end{tabular}
\end{table}

\paragraph{Interpretation.}
Together, the trajectory and endpoint measurements support a regular
depth-conditioned progression that is only mildly sensitive to the tested
changes of intermediate coordinates.

\section{Analysis of the LiFT objective}
\label{sec:app-derivations}

This appendix states the assumptions behind the conditional-regression
result of \cref{sec:lift_method}, derives the training gradient with a
detached anchor, and interprets the reference path through Euclidean
minimum action, Gaussian transport, and Fisher geometry, with Gaussian KL
divergence recovering its fitting objective.

\subsection{Conditional regression result}
\label{sec:app-statistical}

The target $u^\star$ depends on sampled endpoints that are not
individually available at inference, so its regression optimum is
conditional on the predictor's inputs. Fix an initial predictor
$b=b(x_t,t)$ and write $z=(x_t,t)$, including any conditioning variable
when present. Let $v^\star(z)=\mathbb E[u^\star\mid z]$ denote the conditional outer
velocity. For a fixed depth $s$, the reference target and its conditional
mean are
\begin{equation}
  Y_s=(1-s)b(z)+s u^\star,
  \qquad
  m_s(z)=\mathbb{E}[Y_s\mid z]
        =(1-s)b(z)+s v^\star(z).
  \label{eq:app-conditional-target}
\end{equation}
For any deterministic prediction $f(z)$ and finite second moments, the
squared-error decomposition is
\begin{equation}
  \mathbb{E}\!\left[\|f(z)-Y_s\|_2^2\mid z\right]
  =\|f(z)-m_s(z)\|_2^2
   +s^2\mathbb{E}\!\left[\|u^\star-v^\star(z)\|_2^2\mid z\right].
  \label{eq:app-conditional-mse}
\end{equation}
The random depth grid is independent of the endpoint pair, so
conditioning additionally on the full grid $\mathcal S$ leaves
$\mathbb E[u^\star\mid z,\mathcal S]=v^\star(z)$. The same decomposition
therefore applies at each $s_k$, even though a recurrent prediction can
depend on the preceding depth coordinates.
Consequently, the unrestricted regression optimum progresses from the
fixed initial predictor toward the population velocity field, reaching
the ordinary flow-matching optimum at $s=1$. These optima assume a
fixed anchor function and unrestricted predictions; parameter sharing,
finite capacity, and optimization can prevent their simultaneous
realization.

\subsection{Gradient flow with a detached anchor}
\label{sec:app-gradient}

During training, $b$ is recomputed from the current model but detached
within each update. If $J_\theta u_k$ is the total Jacobian of the $k$th
readout through the unrolled computation, the trajectory gradient is
\begin{equation}
  g_\theta
  =\mathbb{E}_{x_0,x_1,t,\mathcal S}\!\left[\frac{2}{Kd}\sum_{k=1}^K
      (J_\theta u_k)^\top(u_k-\bar u_{s_k})\right].
  \label{eq:app-detached-gradient}
\end{equation}
The Jacobian includes dependence on $h_0$ through both initialization and
boundary reinjection, excluding only the target-side dependence through
$b=\operatorname{sg}(u_0)$. Thus, the fixed-anchor path arguments apply
to the reference within each update, while the anchor itself evolves
across updates.

The prelude loss in \cref{eq:lift_training_loss} adds
$\frac{2\lambda}{d}\mathbb E[(J_\theta u_0)^\top(u_0-u^\star)]$
to this gradient, directly supervising how the initial prediction is
learned across updates. Since the anchor remains detached in the
trajectory terms, this supervision leaves the form of the reference path
and its fixed-endpoint derivations unchanged.

\paragraph{Scope of the construction.}
The regression and path arguments characterize the supervision:
conditional regression describes optimal predictions for a fixed anchor,
and the path arguments below characterize the reference connecting the
endpoints. The recurrent core is not required to satisfy a fixed-point
equation or implement a numerical solver for an inner ODE. Consequently, changing
the inference-depth grid tests how the learned recurrence generalizes;
these analyses provide no
guarantee of reduced numerical integration error.

\subsection{Euclidean minimum action}
\label{sec:app-action}
\label{app:minimum-action}

For the path arguments, fix a training example $(x_0,x_1,t)$ and the
endpoints $b=\operatorname{sg}(u_0)$ and $u^\star=x_1-x_0$. Paths have unit
duration and absolutely continuous means with square-integrable
derivatives. The Gaussian constructions restrict every admissible path
to a common, fixed covariance $\sigma^2I$, where $\sigma>0$.
The model fits predictions at the random coordinates
$\mathcal S=(s_0,\ldots,s_K)$ defined in \cref{eq:lift_depth_grid},
using weights $1/K$ and coordinate normalization $1/d$. These regression
weights specify how the selected path is fitted, separately from the
assumptions that select the path itself.

Fix $b,u^\star\in\mathbb R^d$ and let $\Delta=u^\star-b$. For any
absolutely continuous path $\gamma$ with square-integrable derivative and
endpoints $\gamma(0)=b$, $\gamma(1)=u^\star$, the endpoint constraints give
$\int_0^1\dot\gamma(s)\,\mathrm ds=\Delta$. Expanding a squared residual
therefore yields
\begin{equation}
  \label{eq:app-action-decomposition}
  \mathcal A[\gamma]=\frac12\int_0^1\|\dot\gamma(s)\|_2^2\,\mathrm ds
  =\frac12\|\Delta\|_2^2
   +\frac12\int_0^1\|\dot\gamma(s)-\Delta\|_2^2\,\mathrm ds.
\end{equation}
The second term is nonnegative and vanishes if and only if
$\dot\gamma(s)=\Delta$ almost everywhere. Integrating with the initial
boundary condition gives the unique minimizer
$\gamma(s)=b+s\Delta=(1-s)b+s u^\star$.\footnote{The same straight
path minimizes the action under any constant positive-definite metric;
a prediction-dependent metric can instead yield a curved path.}

The same argument applies to a nonuniform depth grid. Let
$\delta_k=s_k-s_{k-1}>0$, $z_0=b$, $z_K=u^\star$, and interpolate the
knots linearly between successive coordinates. The resulting action is
\begin{equation}
  \begin{aligned}
    \mathcal A_{\mathcal S}(z_{0:K})
    &=\frac12\sum_{k=1}^K
       \frac{\|z_k-z_{k-1}\|_2^2}{\delta_k}\\
    &=\frac12\|\Delta\|_2^2+
      \frac12\sum_{k=1}^K\frac{
        \|z_k-z_{k-1}-\delta_k\Delta\|_2^2}{\delta_k}.
  \end{aligned}
  \label{eq:app-discrete-action}
\end{equation}
Since $\sum_k\delta_k=1$, the minimizing increments are
$z_k-z_{k-1}=\delta_k\Delta$, giving
$z_k=b+s_k\Delta=\bar u_{s_k}$ for every sampled grid. Uniform grids
recover increments $\Delta/K$ as a special case. Thus randomizing the
coordinates changes the reference knots without changing the underlying
minimum-action path.

The factors $1/\delta_k$ belong to the action of the interpolated path
and thereby determine its minimizing knots. Once these knots are
selected as targets, \cref{eq:lift_trajectory_loss} averages
their prediction errors with weights $1/K$. Applying the action directly
to the model's successive increments would instead define a different
loss and produce different gradients.

\subsection{Gaussian, Fisher, and KL interpretations}
\label{sec:app-geometric}

\subsubsection{Gaussian probability transport}
\label{sec:app-transport}

A probabilistic view lifts a path of means $\mu_s$ into the translated
Gaussian family
\begin{equation}
  q_s(w)=\mathcal{N}(w;\mu_s,\sigma^2 I),
  \qquad \mu_0=b,\quad\mu_1=u^\star.
  \label{eq:app-gaussian-path}
\end{equation}
Here $w\in\mathbb{R}^d$ is a velocity-space variable, and the auxiliary
distributions describe the fixed training example. The translation field
$a_s(w)=\dot\mu_s$ transports this family and satisfies
\begin{equation}
  \partial_s q_s(w)+\nabla_w\!\cdot\!\left(q_s(w)a_s(w)\right)=0,
  \qquad
  \mathcal{A}_{\mathrm{prob}}[q,a]
   =\frac{1}{2}\int_0^1\!\int_{\mathbb{R}^d}
     \|a_s(w)\|_2^2 q_s(w)\,\mathrm{d}w\,\mathrm{d}s.
  \label{eq:app-probability-action}
\end{equation}
Any admissible field transporting this density path satisfies
$\mathbb{E}_{q_s}[a_s(w)]=\dot\mu_s$, assuming
finite energy and sufficient decay for integration by parts. Jensen's
inequality then gives
$\mathbb{E}_{q_s}\|a_s(w)\|_2^2\geq\|\dot\mu_s\|_2^2$, with equality
for the constant field $a_s=\dot\mu_s$. Minimizing the transport action
within this fixed-covariance family thus reduces to minimizing
$\tfrac12\int_0^1\|\dot\mu_s\|_2^2\,\mathrm{d}s$. By the preceding
derivation, the reference distribution is
\begin{equation}
  \bar q_s=\mathcal{N}\!\left((1-s)b+s u^\star,\sigma^2 I\right).
  \label{eq:app-reference-distribution}
\end{equation}

Associating each readout with
$q_{\theta,k}=\mathcal{N}(u_k,\sigma^2 I)$ and fitting it to
$\bar q_{s_k}$ by KL divergence gives the trajectory objective, as
derived in \cref{sec:app-kl}. In this interpretation, the auxiliary
Gaussians are transported by shifting their means while keeping covariance
fixed across depth. The deterministic network passes the terminal mean
to the outer sampler as its velocity prediction; it neither samples $w$
nor estimates posterior uncertainty.

\subsubsection{Fisher geometry with fixed covariance}
\label{sec:app-fisher}

The fixed-covariance family also gives the reference an
information-geometric interpretation. For
$q(w;\mu)=\mathcal{N}(w;\mu,\sigma^2 I)$,
the score with respect to the mean and its Fisher information matrix are
\begin{equation}
  \nabla_\mu\log q(w;\mu)=\sigma^{-2}(w-\mu),
  \qquad
  G_F(\mu)=\mathbb{E}_{q}
    \left[(\nabla_\mu\log q)(\nabla_\mu\log q)^\top\right]
    =\sigma^{-2}I.
  \label{eq:app-fisher-metric}
\end{equation}
The Fisher metric on this mean submanifold is constant, so its path energy
is a scaled Euclidean action:
\begin{equation}
  \mathcal{A}_F[\mu]
   =\frac{1}{2}\int_0^1\dot\mu_s^\top G_F\dot\mu_s\,\mathrm{d}s
   =\sigma^{-2}\mathcal{A}[\mu].
  \label{eq:app-fisher-action}
\end{equation}
The unique constant-speed energy minimizer between the specified endpoints
is therefore $\mu_s=\bar u_s$. Fitting the predicted Gaussian means to
this geodesic by KL again gives \cref{eq:lift_trajectory_loss}.

This conclusion requires covariance to be fixed along every admissible
path. Equal endpoint covariances alone are insufficient, because allowing
covariance to vary changes the manifold and can give geodesics that leave
this subfamily~\citep{nielsen2023fisher}. A common, constant positive-definite
covariance $\Sigma$ would still give a straight mean geodesic under
$G_F=\Sigma^{-1}$, but its KL fitting criterion would be a Mahalanobis
error. The isotropic assumption makes that error proportional to the
unweighted MSE used here.

\subsubsection{Gaussian KL projection}
\label{sec:app-kl}

Once the reference path is specified, Gaussian KL provides a direct
derivation of its fitting loss. At a depth $s_k$, let
$\bar q_k=\mathcal{N}(\bar u_{s_k},\sigma^2 I)$ and
$q_{\theta,k}=\mathcal{N}(u_k,\sigma^2 I)$. Their log-density ratio gives
\begin{equation}
  \begin{aligned}
    D_{\mathrm{KL}}(\bar q_k\|q_{\theta,k})
    &=\frac{1}{2\sigma^2}\mathbb{E}_{w\sim\bar q_k}
       \left[\|w-u_k\|_2^2-\|w-\bar u_{s_k}\|_2^2\right]\\
    &=\frac{1}{2\sigma^2}\|u_k-\bar u_{s_k}\|_2^2.
  \end{aligned}
  \label{eq:app-gaussian-kl}
\end{equation}
The cross term vanishes because
$\mathbb{E}_{\bar q_k}[w-\bar u_{s_k}]=0$. Thus, after averaging over
training examples and recurrent depths,
\begin{equation}
  \mathcal{L}_{\mathrm{KL}}
  =\mathbb{E}_{x_0,x_1,t,\mathcal S}\!\left[
    \frac{1}{K}\sum_{k=1}^K
    D_{\mathrm{KL}}(\bar q_k\|q_{\theta,k})\right]
  =\frac{d}{2\sigma^2}\mathcal{L}_{\mathrm{LiFT}}.
  \label{eq:app-kl-equivalence}
\end{equation}
The factor $d$ accounts for coordinate averaging in the implemented MSE.
Because $\sigma$ is fixed, the objectives differ only by a positive
constant and have proportional gradients under the same
detached-boundary convention. Their equivalence therefore requires no
Gaussian sampling or variance parameter in the implementation. Whereas
the action and geometric arguments select a reference path, KL recovers
the criterion used to fit it once that path is specified.

\FloatBarrier
\section{Reproducibility and implementation}
\label{app:experimental-details}

The main comparisons share the same data pipeline, transformer
block design, and optimization settings, allowing dense and
recurrent architectures to be evaluated under a common protocol.
\cref{app:full_results} provides the numerical results for
these comparisons.

\subsection{Data and architecture}
\label{app:data_and_models}

We use class-conditional ImageNet-1k at $256\times256$ resolution.
Images follow the ADM/DiT preprocessing convention: BOX
downsampling, bicubic resizing, and a $256\times256$ center crop.
Each preprocessed image is encoded into one cached SD-VAE posterior
sample, with posterior noise drawn on CPU from a seed determined by
the cache seed and global image index. The resulting
$4\times32\times32$ latents are scaled by $0.18215$ and stored in
FP16. Training reuses these cached representations while sampling
fresh flow noise and generative times.

Dense DiT and LiFT use the same SpeedrunDiT-adapted transformer
blocks, with RMSNorm, per-head query/key normalization,
two-dimensional rotary embeddings, GELU MLPs, and
adaLN-zero-style conditioning with gated residual branches.
\cref{tab:backbone_dimensions} gives the backbone dimensions;
all models use $2\times2$ latent patches, yielding 256 tokens
per image.

\begin{table}[htbp]
    \centering
    \small
    \caption{
        Backbone dimensions shared by the dense and recurrent
        models. Dense depth counts transformer blocks.
    }
    \label{tab:backbone_dimensions}
    \begin{tabular}{lrrr}
        \toprule
        Backbone & Width & Attention heads & Dense depth \\
        \midrule
        B/2  & 768  & 12 & 12 \\
        L/2  & 1024 & 16 & 24 \\
        XL/2 & 1152 & 16 & 28 \\
        \bottomrule
    \end{tabular}
\end{table}

Within this shared block design, LiFT uses four prelude blocks and
no transformer blocks in the coda for the main comparisons. A
shared prediction head reads out the prelude and recurrent states
during training, whereas sampling requires only the final readout.
Because the core is shared, each LiFT model has fewer parameters than
its dense backbone. The depth embedding adds a small overhead relative
to a dense model with the same number of unique blocks, and the
parameter count remains fixed when the loop count changes. \cref{tab:all_model_configurations}
collects the configurations and their analytic training costs.

\begin{table}[tbp]
\centering
\small
\setlength{\tabcolsep}{5pt}
\caption{Complete model configurations and training resources for the
checkpoints in \cref{app:all_evaluations}, with R denoting the shared-core
block count. All LiFT models use four prelude blocks and no transformer
coda, giving $L_{\mathrm{unique}}=4+L_R$ unique blocks and
$L_{\mathrm{exec}}^{\mathrm{train}}=4+K_{\mathrm{train}}L_R$ executed blocks
per training forward pass. Executed depth matches the dense reference
within each backbone scale, with its shared value centered across each
LiFT group. Parameter counts $N$ are obtained from the implementation and
include conditioning and readout parameters, whereas block counts omit
this work. The analytic forward-plus-backward estimate $C_{\mathrm{train}}$
accounts for it using the protocol in \cref{app:training_protocol} and
the accounting in \cref{app:compute_accounting}, and is reported in EFLOPs
($10^{18}$ FLOPs).}
\label{tab:all_model_configurations}
\label{tab:checkpoint_resources}
\begin{tabular}{lrrrrrr}
\toprule
Model & $L_R$ & $K_{\mathrm{train}}$ & $L_{\mathrm{unique}}$
& $L_{\mathrm{exec}}^{\mathrm{train}}$ & $N$ (M)
& $C_{\mathrm{train}}$ (EFLOPs) \\
\midrule
\multicolumn{7}{l}{\textit{Dense references}} \\
DiT-B/2 & -- & -- & 12 & 12 & 130.30 & 17.666 \\
DiT-L/2 & -- & -- & 24 & 24 & 457.83 & 61.973 \\
DiT-XL/2 & -- & -- & 28 & 28 & 674.82 & 91.098 \\
\midrule
\multicolumn{7}{l}{\textit{LiFT configurations}} \\
LiFT-B/2-R1 & 1 & 8 & 5 & \multirow{3}{*}{12} & 56.69 & 17.697 \\
LiFT-B/2-R2 & 2 & 4 & 6 &  & 67.32 & 17.681 \\
LiFT-B/2-R4 & 4 & 2 & 8 &  & 88.58 & 17.673 \\
\midrule
LiFT-L/2-R1 & 1 & 20 & 5 & \multirow{5}{*}{24} & 100.23 & 62.089 \\
LiFT-L/2-R2 & 2 & 10 & 6 &  & 119.12 & 62.031 \\
LiFT-L/2-R4 & 4 & 5 & 8 &  & 156.90 & 62.002 \\
LiFT-L/2-R5 & 5 & 4 & 9 &  & 175.79 & 61.996 \\
LiFT-L/2-R10 & 10 & 2 & 14 &  & 270.24 & 61.984 \\
\midrule
LiFT-XL/2-R1 & 1 & 24 & 5 & \multirow{7}{*}{28} & 126.62 & 91.263 \\
LiFT-XL/2-R2 & 2 & 12 & 6 &  & 150.53 & 91.181 \\
LiFT-XL/2-R3 & 3 & 8 & 7 &  & 174.43 & 91.153 \\
LiFT-XL/2-R4 & 4 & 6 & 8 &  & 198.34 & 91.139 \\
LiFT-XL/2-R6 & 6 & 4 & 10 &  & 246.15 & 91.125 \\
LiFT-XL/2-R8 & 8 & 3 & 12 &  & 293.96 & 91.118 \\
LiFT-XL/2-R12 & 12 & 2 & 16 &  & 389.58 & 91.111 \\
\bottomrule
\end{tabular}
\end{table}

\subsection{Training protocol}
\label{app:training_protocol}
\label{app:checkpoint_resources}

The main protocol uses 500,000 optimizer updates with a global
batch size of 256 and the optimization and numerical settings in
\cref{tab:optimization_settings}. Under this common protocol,
dense models use conditional flow matching, while LiFT uses the
trajectory objective with an auxiliary prelude loss of weight
$\lambda=0.01$, added once outside the average over recurrent
predictions.

\begin{table}[htbp]
    \centering
    \small
    \caption{Shared settings for the main training protocol.}
    \label{tab:optimization_settings}
    \begin{tabular}{ll}
        \toprule
        Setting & Value \\
        \midrule
        Optimizer & AdamW \\
        Training updates & 500,000 \\
        Global batch size & 256 \\
        Learning rate & $10^{-4}$ \\
        Schedule & 500 warmup updates, then constant \\
        Adam coefficients & $(0.9,\,0.999)$ \\
        Adam $\epsilon$ & $10^{-8}$ \\
        Weight decay & $0$ \\
        Gradient clipping & Global norm $1$ \\
        EMA decay & $0.999$ \\
        Training seed & $42$ \\
        Computation precision & BF16 \\
        Parameter and loss-accumulation precision & FP32 \\
        \bottomrule
    \end{tabular}
\end{table}

Both model families use standard Gaussian flow noise, uniform
generative times, and a linear noise-to-data interpolant, retaining
class labels throughout training without label dropout. LiFT also
samples interior depth coordinates independently for each example
and update, then sorts them and appends the endpoint $s=1$. The
number of recurrent applications remains fixed within each run.

The reported B/2, L/2, and XL/2 checkpoints in
\cref{app:all_evaluations} were trained on four NVIDIA H100 GPUs,
with 64 examples per GPU and no gradient accumulation;
\cref{tab:all_model_configurations} reports their analytic training
costs, computed as in \cref{app:compute_accounting}.

\subsection{Training and sampling pseudocode}
\label{app:pseudocode}

In each training update, \cref{alg:lift-training} independently
samples the noise, generative time, and depth coordinates for every
example in the minibatch. Each $\operatorname{MSE}$ averages over examples
and latent coordinates, while the loop average is written explicitly.
The prelude weight $\lambda$ controls direct supervision of the initial
prediction, using the experimental value given in
\cref{app:training_protocol}. Only the copy of that prediction
used in the trajectory target is detached; the auxiliary loss
differentiates through the prediction itself.

\begin{algorithm}[htbp]
\caption{LiFT trajectory training with random depth coordinates}
\label{alg:lift-training}
\begin{algorithmic}[1]
\REQUIRE Parameters $\theta$, training depth $K\geq1$, minibatch $(x_1,y)$, prelude weight $\lambda$
\STATE Sample $x_0\sim\mathcal N(0,I)$ and $t\sim\mathcal U(0,1)$ per example
\STATE $x_t\gets(1-t)x_0+t x_1$; $u^\star\gets x_1-x_0$
\STATE Draw $K-1$ independent uniform depths and sort them into $s_1,\ldots,s_{K-1}$
\STATE Set $s_0\gets0$, $s_K\gets1$ \COMMENT{No interior samples when $K=1$}
\STATE $c\gets e_t(t)+e_y(y)$; $h_0\gets P_\theta(x_t,c)$; $h\gets h_0$
\STATE $u_0\gets \mathcal D_\theta(h_0,c)$; $b\gets\operatorname{sg}(u_0)$
\STATE $\ell\gets\lambda\operatorname{MSE}(u_0,u^\star)$ \COMMENT{Outside the loop average}
\FOR{$k=1,\ldots,K$}
  \STATE $h\gets R_\theta\!\left(\operatorname{RMSNorm}(h)+h_0,\ c+e_s(s_k)\right)$
  \STATE $u_k\gets \mathcal D_\theta(h,c)$
  \STATE $\bar u_k\gets(1-s_k)b+s_k u^\star$
  \STATE $\ell\gets\ell+\operatorname{MSE}(u_k,\bar u_k)/K$
\ENDFOR
\STATE Backpropagate $\ell$ through all predictions and recurrent hidden states
\STATE Update $\theta$ with the optimizer; update the EMA parameters
\end{algorithmic}
\end{algorithm}

To preserve numerical ordering and endpoint spacing, the implementation
draws and sorts interior coordinates in FP64, then applies
$s_i=(1-K\epsilon)r_{(i)}+i\epsilon$, with $\epsilon=10^{-6}$, before
casting the grid to FP32. Here $r_{(i)}$ is the $i$th sorted uniform draw,
and the endpoints remain exactly zero and one. The resulting separation
leaves the loss weights unchanged and introduces no step-size factor
into the core.

\begin{algorithm}[htbp]
\caption{LiFT sampling with $T$ Euler steps and $K_{\mathrm{inf}}$ inner loops}
\label{alg:lift-sampling}
\begin{algorithmic}[1]
\REQUIRE EMA parameters, condition $y$, integration steps $T$, inner loops $K_{\mathrm{inf}}$
\STATE Sample $x\sim\mathcal N(0,I)$
\FOR{$j=0,\ldots,T-1$}
  \STATE $t\gets j/T$; $c\gets e_t(t)+e_y(y)$
  \STATE $h_0\gets P_\theta(x,c)$; $h\gets h_0$
  \FOR{$k=1,\ldots,K_{\mathrm{inf}}$}
    \STATE $s\gets k/K_{\mathrm{inf}}$
    \STATE $h\gets R_\theta\!\left(\operatorname{RMSNorm}(h)+h_0,\ c+e_s(s)\right)$
  \ENDFOR
  \STATE $x\gets x+\mathcal D_\theta(h,c)/T$
\ENDFOR
\RETURN Decode $x$ with the latent tokenizer's decoder
\end{algorithmic}
\end{algorithm}

At inference, only the final prediction advances the outer sample, so
the sampling algorithm omits intermediate readouts and the training
anchor. It uses no classifier-free guidance (scale 1), as specified in
the evaluation protocol.

\FloatBarrier
\subsection{Evaluation protocol}
\label{app:evaluation_protocol}

Each setting generates 50,000 images, with 50 samples per class.
Sample $i\in\{0,\ldots,49{,}999\}$ receives class label
$i\bmod1000$ and initial Gaussian noise generated in FP32 on
CPU with seed $200042+i$. This assignment pairs the initial noise
and class labels across models and sampling budgets. We preserve
these sample identities across distributed generation and resumed
jobs to maintain the pairing throughout evaluation.

Sampling uses EMA weights and Euler integration from noise to
data, without classifier-free guidance (scale 1). Each integration step
requires one velocity evaluation. Depth sweeps hold the number of
steps at 50, while step sweeps vary it.
At every integration step, LiFT reinitializes the
prelude state and uses the uniform inference-depth coordinates
$s_k=k/K_{\mathrm{inf}}$.

Transformer inference uses BF16, while Euler updates and VAE
decoding use FP32. We decode samples with
\texttt{stabilityai/sd-vae-ft-mse}, using a latent scaling factor
of $0.18215$.

The decoded samples are evaluated with the official ADM Inception
evaluator. FID and sFID compare features from all 50,000 generated
images with the published ImageNet reference statistics, while IS
is computed from the same generated images.

\subsection{Compute accounting}
\label{app:compute_accounting}

The $32\times32$ latent grid contains
$n=(32/2)^2=256$ patch tokens. With $U=500{,}000$ updates and
global batch size $B_{\mathrm{batch}}=256$, the training-data
budget is
\begin{equation}
    D=U B_{\mathrm{batch}}\, n
     =3.2768\times10^{10}
\end{equation}
latent-patch tokens, corresponding to 128 million image
presentations. This data budget counts repeated image
presentations, with recurrent computation over the same tokens
accounted for separately below. Parameter counts include all
trainable embedding, conditioning, transformer, and readout
parameters.

To account for computation, we estimate FLOPs from the architecture,
counting two operations per multiply-add. Let
$C_{\mathrm{forward}}^{\mathrm{train}}(K_{\mathrm{train}})$
denote the forward cost per example during training. For LiFT,
this includes the prelude, all recurrent applications, depth
conditioning, and the anchor and intermediate readouts.
For the dense baseline, it includes the ordinary forward pass
and final prediction. Total training computation is estimated as
\begin{equation}
    C_{\mathrm{train}}
    \approx
    3 U B_{\mathrm{batch}}\,
    C_{\mathrm{forward}}^{\mathrm{train}}(K_{\mathrm{train}}),
    \label{eq:train_compute}
\end{equation}
with the factor of three approximating the combined forward
and backward cost. The depth argument is omitted for dense models.
\Cref{app:trajectory_compute} expands both costs in terms of the
individual model components.

Sampling requires only the final readout. Its cost per generated
image is therefore
\begin{equation}
    C_{\mathrm{infer}}
    =
    T\,
    C_{\mathrm{forward}}(K_{\mathrm{inf}}),
    \label{eq:infer_compute}
\end{equation}
where $C_{\mathrm{forward}}$ includes patch embedding, the prelude,
every core application, conditioning, and the final prediction
head, but excludes SD-VAE decoding.

Within each backbone scale, matching executed transformer depth
approximately matches training computation; the following derivation
quantifies the remaining overhead from conditioning and readouts when
the coda contains no transformer blocks.

\subsubsection{Cost of trajectory supervision}
\label{app:trajectory_compute}

Because trajectory supervision reads out every recurrent state, each
operation in the prediction branch contributes repeatedly to the
training cost. A zero-block coda, paired with the four-block prelude
used in our comparisons, limits this repeated computation to the
prediction head. We quantify its cost using the analytic accounting of
\cref{app:compute_accounting}, with two operations per multiply-add.

\paragraph{General training and sampling costs.}
Let $L_P$, $L_R$, and $L_D$ denote the numbers of transformer blocks
in the prelude, shared core, and prediction-only coda, respectively.
Write $f_{\mathrm{blk}}$ for the forward cost of one block,
$f_{\mathrm{head}}$ for the prediction head after the coda,
$f_s$ for one depth embedding, and $f_{\mathrm{io}}$ for the patch
embedding and outer-time conditioning computed once per example.
With $K=K_{\mathrm{train}}$, training computes $u_0,\ldots,u_K$,
giving
\begin{equation}
    \begin{aligned}
    C_{\mathrm{forward}}^{\mathrm{train}}(K)
    ={}& f_{\mathrm{io}}+Kf_s
    +\bigl[L_P+K L_R+(K+1)L_D\bigr]f_{\mathrm{blk}}\\
    &+(K+1)f_{\mathrm{head}}.
    \end{aligned}
    \label{eq:trajectory_forward_cost}
\end{equation}
The $K+1$ prediction branches include the initial estimate used as
the anchor and every recurrent readout. Sampling requires
only the final prediction, reducing the cost at depth
$K=K_{\mathrm{inf}}$ to
\begin{equation}
    C_{\mathrm{forward}}(K)
    =f_{\mathrm{io}}+Kf_s
    +(L_P+K L_R+L_D)f_{\mathrm{blk}}+f_{\mathrm{head}}.
    \label{eq:trajectory_sampling_cost}
\end{equation}
Although all recurrent applications share the same parameters,
each application contributes to the computation in these expressions.

Because the stop-gradient on $b=\operatorname{sg}(u_0)$ removes only the
target-side derivative, backpropagation still traverses the prelude state
and every recurrent transition, and the auxiliary prelude loss also
differentiates through $u_0$. The usual approximation of the backward pass
as twice the forward cost therefore applies, and substituting these
forward costs into \cref{eq:train_compute,eq:infer_compute} gives the
total training and inference budgets. The forward cost includes the anchor computation
whether or not the auxiliary prelude loss is enabled.

\paragraph{Why we use a zero-block coda.}
Consider a dense model with the same block design and width, and
$L=L_P+K L_R+L_D$ blocks. Its forward cost is
$C_{\mathrm{forward}}^{\mathrm{dense}}
=f_{\mathrm{io}}+L f_{\mathrm{blk}}+f_{\mathrm{head}}$.
This dense model matches LiFT's executed transformer depth at sampling
depth $K$, but LiFT's trajectory training incurs an additional
\begin{equation}
    C_{\mathrm{forward}}^{\mathrm{train}}(K)
    -C_{\mathrm{forward}}^{\mathrm{dense}}
    =K\bigl(L_D f_{\mathrm{blk}}+f_{\mathrm{head}}+f_s\bigr).
    \label{eq:trajectory_cost_overhead}
\end{equation}
The repeated coda evaluations account for the transformer-block
overhead that remains when only sampling depth is matched. We
therefore choose $L_D=0$ and
$L_P+K_{\mathrm{train}}L_R=L$ in the matched-training comparisons.
This choice equalizes transformer-block computation, leaving the
additional prediction heads and depth embeddings as the remaining
overhead. Under the same training-cost approximation, their relative
cost is
\begin{equation}
    \frac{C_{\mathrm{train}}^{\mathrm{LiFT}}}
         {C_{\mathrm{train}}^{\mathrm{dense}}}-1
    =\frac{K_{\mathrm{train}}(f_{\mathrm{head}}+f_s)}
           {C_{\mathrm{forward}}^{\mathrm{dense}}},
    \label{eq:zero_coda_overhead}
\end{equation}
for equal update and global-batch budgets. The total FLOPs are
therefore approximately, rather than exactly, matched.

\paragraph{Numerical accounting for our models.}
For $n$ patch tokens, hidden width $w$, MLP width $m$, patch
dimension $q$, and sinusoidal embedding dimension $f$, our
implementation counts
\begin{equation}
    \begin{alignedat}{2}
        f_{\mathrm{blk}} &= n(8w^2+4wm+4nw)+12w^2,
        &\qquad f_s &= 2(fw+w^2),\\[2pt]
        f_{\mathrm{io}} &= 2nqw+2(fw+w^2),
        &\qquad f_{\mathrm{head}} &= 4w^2+2nwq.
    \end{alignedat}
    \label{eq:trajectory_component_flops}
\end{equation}
The block expression includes attention projections, both attention
matrix products, the MLP, and the conditioning projection.
For our SD-VAE models, $n=256$, $q=2^2\!\times4=16$, $m=4w$,
and $f=256$. Including all supervised readouts gives the
representative comparisons in \cref{tab:trajectory_compute}.

\begin{table}[H]
    \centering
    \small
    \setlength{\tabcolsep}{5pt}
    \caption{Analytic training compute for representative matched-depth
    configurations. Each model uses 500,000 updates and a global batch
    size of 256; one EFLOP is $10^{18}$ operations. LiFT uses $L_P=4$
    and $L_D=0$. The totals include the forward and approximate backward
    cost of conditioning and all trajectory readouts. Overheads are
    computed before rounding the totals.}
    \label{tab:trajectory_compute}
    \begin{tabular}{lrrrrr}
        \toprule
        Backbone & $L_R$ & $K_{\mathrm{train}}$
        & Dense (EFLOPs) & LiFT (EFLOPs) & Extra cost \\
        \midrule
        B/2  & 2  & 4 & 17.6655 & 17.6812 & $0.0889\%$ \\
        L/2  & 10 & 2 & 61.9726 & 61.9842 & $0.0188\%$ \\
        XL/2 & 4  & 6 & 91.0976 & 91.1391 & $0.0455\%$ \\
        \bottomrule
    \end{tabular}
\end{table}

Across all matched-depth configurations in \cref{tab:all_model_configurations},
the additional cost ranges from $0.044\%$ to $0.178\%$ for B/2,
from $0.019\%$ to $0.188\%$ for L/2, and from $0.015\%$ to
$0.182\%$ for XL/2. Thus, every configuration differs from its
same-scale dense reference by less than $0.2\%$ under this accounting.

\end{document}